\PassOptionsToPackage{table}{xcolor}
\documentclass[11pt]{article}
\usepackage[final]{acl}
\usepackage{times}
\usepackage{multirow}
\usepackage{latexsym}
\usepackage[T1]{fontenc}
\usepackage[utf8]{inputenc}
\usepackage{microtype}
\usepackage{inconsolata}
\usepackage{amsmath}
\usepackage{amssymb}
\usepackage{booktabs}
\usepackage{enumitem}
\usepackage{graphicx}
\usepackage{listings}
\usepackage{tikz}
\usepackage{pgfplots}
\pgfplotsset{compat=1.18}
\usepackage{xcolor}
\usepackage{fancyhdr}
\usepackage{lastpage}
\usepackage{multirow}

\setlist[itemize]{noitemsep,topsep=2pt,leftmargin=*}
\setlist[enumerate]{noitemsep,topsep=2pt,leftmargin=*}

\makeatletter
\g@addto@macro{\UrlBreaks}{\UrlOrds}
\makeatother

\usetikzlibrary{positioning, arrows.meta, fit, backgrounds, shapes.geometric, calc}

\title{AILS-NTUA at SemEval-2026 Task 10: Agentic LLMs for Psycholinguistic Marker Extraction and Conspiracy Endorsement Detection}

\author{Panagiotis Alexios Spanakis \hspace{0.8em} Maria Lymperaiou \hspace{0.8em} Giorgos Filandrianos \\
\textbf{Athanasios Voulodimos \quad Giorgos Stamou} \\
School of Electrical and Computer Engineering, AILS Laboratory \\
National Technical University of Athens \\
\href{mailto:spanakis01@gmail.com}{\texttt{spanakis01@gmail.com}}, \href{mailto:marialymp@ails.ece.ntua.gr}{\texttt{\{marialymp,}} \href{mailto:geofila@ails.ece.ntua.gr}{\texttt{geofila\}@ails.ece.ntua.gr}} \\
\href{mailto:thanosv@mail.ntua.gr}{\texttt{thanosv@mail.ntua.gr}}, \href{mailto:gstam@cs.ntua.gr}{\texttt{gstam@cs.ntua.gr}}}

\begin{document}
\maketitle
    

\begin{abstract}
	This paper presents a novel agentic LLM pipeline for SemEval-2026 Task 10 that jointly extracts psycholinguistic conspiracy markers and detects conspiracy endorsement.
	Unlike traditional classifiers that conflate semantic reasoning with structural localization, our decoupled design isolates these challenges.
	For marker extraction, we propose Dynamic Discriminative Chain-of-Thought (DD-CoT) with deterministic anchoring to resolve semantic ambiguity and character-level brittleness.
	For conspiracy detection, an ``Anti-Echo Chamber'' architecture, consisting of an adversarial Parallel Council adjudicated by a Calibrated Judge, overcomes the ``Reporter Trap,'' where models falsely penalize objective reporting.
	Achieving 0.24 Macro F1 (+100\% over baseline) on S1 and 0.79 Macro F1 (+49\%) on S2, with the S1 system ranking 3rd on the development leaderboard, our approach establishes a versatile paradigm for interpretable, psycholinguistically-grounded NLP.
\end{abstract}

\section{Introduction}
Humans have long exhibited a tendency to endorse conspiracy theories,
particularly in contexts of uncertainty, threat, and social upheaval. Such
beliefs are created and distributed to support human need for addressing
existential or social issues and strengthen their sense of identity and
belonging \cite{psychology-of-conspiracy, belief}. Despite their psychological
appeal, conspiracies are associated with harmful consequences, limiting trust
in well-documented facts and public decisions, while exacerbating political
polarization and misinformation patterns \cite{understanding}.

The rise of AI strengthened the link between conspiracy identification and language, the primary medium through which conspiracies are articulated and disseminated. Conspiratorial statements are often subtly embedded in linguistic strategies that evoke emotion and attribute agency \cite{miani2022loco, rains2023psycholinguistic}, indicating that effective detection extends beyond superficial textual cues.

Large Language Models (LLMs) have revolutionized linguistic research, enabling
deep pattern identification and discrimination among their numerous abilities.
However, LLMs have been found to be significantly prone to cognitive biases
\cite{filandrianos-etal-2025-bias} and manipulation via persuasive language
\cite{xu-etal-2024-earth}, while they generate and amplify misinformation
\cite{chen2024combatingmisinformation}. Going one step further,
state-of-the-art LLMs are even able to persuade people to adopt conspiratorial
beliefs to a comparable degree as they can mitigate conspiracy dissemination
\cite{costello2026largelanguagemodelseffectively}, exposing the double-edged
nature of LLMs in the context of factual verification.

The core challenges that conspiratorial discourse poses call for fine-grained
data approaches that allow delving into the linguistic mechanisms that
characterize conspiratorial utterances in an interpretable way. Nevertheless,
prior datasets \cite{shahsavari2020conspiracy, langguth2023coco} frame
conspiracy detection as a coarse-grained classification task, abstracting away
from the particularities of conspiratorial discourse, thus obscuring how
conspiratorial reasoning is formed and expressed in language. To fill this gap,
the \textbf{SemEval 2026 Task 10: Psycholinguistic Conspiracy Marker Extraction
    and Detection} \cite{samory-etal-2026-semeval} emphasizes the localization and categorization of linguistic
markers that signal conspiratorial thinking, complementing detection with
psychologically-backed annotations.

To address the dual challenge of accurate detection and interpretable marker
extraction, models must be capable of capturing both global conspiratorial
intent and fine-grained psycholinguistic cues embedded in language. In our
approach, we leverage LLMs within agentic structures to advance the recognition
of conspiratorial thought: we propose the \textbf{Dynamic
	Discriminative Chain-of-Thought (DD-CoT)} framework which extends the adaptive
nature of Dynamic CoT \cite{ma2024dynamic} to perform semantic discrimination
with deterministic verification for precise marker extraction and an
\textbf{``Anti-Echo Chamber'' council} of contrasting perspectives to separate
conspiracy endorsement from neutral reporting. To the best of our knowledge, our approach constitutes the \textit{first agentic LLM-based method} to combat conspiracy
detection and identification of psycholinguistic features in language.
In short, our contributions are the following:
\begin{itemize}
	\item We introduce the first agentic LLM-based method for psycholinguistic conspiracy
	      marker extraction and endorsement detection.
	\item We propose \textbf{Dynamic Discriminative Chain-of-Thought (DD-CoT)},
	      forcing explicit counter-arguments to deduce semantic ambiguity.
	\item We propose a hybrid extraction architecture decoupling semantic LLM reasoning
	      from deterministic span localization for highly reliable character-accurate outputs.
	\item We provide comprehensive empirical analysis including juror ablation studies,
	      latency profiling, and transferability to 8B open-weights models.
\end{itemize}


Our system ranked 3rd on the S1 development set and 10th on the test set (within 0.05~F1 of the top system), with ablation studies confirming the contribution of each architectural component; high-context irony and implicit stance remain the primary open challenges.
The code for our system is available on GitHub\footnote{\url{https://github.com/panos-span/PsyChoMark_Semeval}}.

\section{Background}
\paragraph{Task description}
The dataset comprises 4,800 annotations spanning 4,100 unique Reddit submission
statements from $>$190 subreddits, divided in two subtasks: \textbf{\textit{i)}
    S1: Conspiracy Marker Extraction} contains textual spans that express core
conspiracy markers grounded in evolutionary psychology. One or more marker
types may appear in each comment, falling in the following categories:
\textsc{Actor} (mentions of individual or group agents), \textsc{Action}
(descriptions of what the actor is doing), \textsc{Effect} (consequences of the
actions), \textsc{Victim} (who is being harmed), \textsc{Evidence} (claims or
proof used to support the theory). \textbf{\textit{ii)} S2: Conspiracy
    Detection} assigns conspiracy-related or not conspiracy-related labels to
Reddit comments. More details about the dataset are provided in App.
\ref{sec:eda}.

\paragraph{Related work}
Early works on NLP conspiratorial discourse on Reddit introduced narrative
motifs correlated with conspiratorial evidence \cite{online-discussions} and
demonstrated that conspiratorial thinking manifests through detectable
psycholinguistic signals in user language \cite{Klein2019Pathways}, with
consequent literature revealing that conspiracy theories exhibit distinctive
narrative frameworks that can be computationally extracted from text
\cite{Tangherlini2020Automated}. These foundational works empowered the
operationalization of conspiracy identification as a classification task,
exemplified by datasets such as COCO \cite{langguth2023coco} and YouNICon
\cite{YouNICon}. Conspiracy detection involves techniques that explicitly model
psycholinguistic signals, such as affective tone, attributional cues and
explanatory framing, in order to provide explanatory evidence of conspiracy
presence in language \cite{rains2023psycholinguistic,
    language-of-conspiracy-theories, marino-etal-2025-linguistic}. The strong
contextualization that LLMs offer inspired the introduction of related
approaches; leveraging appropriate prompting enables accurate multi-label
conspiracy classification, eliminating training demands
\cite{peskine-etal-2023-definitions}. Complementarily, ConspEmoLLM
\cite{liu2024conspemollm} involves emotion-aware LLM fine-tuning on several
conspiratorial tasks, improving detection by leveraging affective signals, with
subsequent extensions focusing on robustness to stylistic and emotional
variation \cite{liu2025conspemollmv2}. Recent evaluations indicate that
LLM-based conspiracy detection often relies on topical shortcuts and struggles
with narrative ambiguity, underscoring the need for approaches grounded in
interpretable psycholinguistic markers \cite{pustet-etal-2024-detection,
    classifying}.

\section{System Overview}

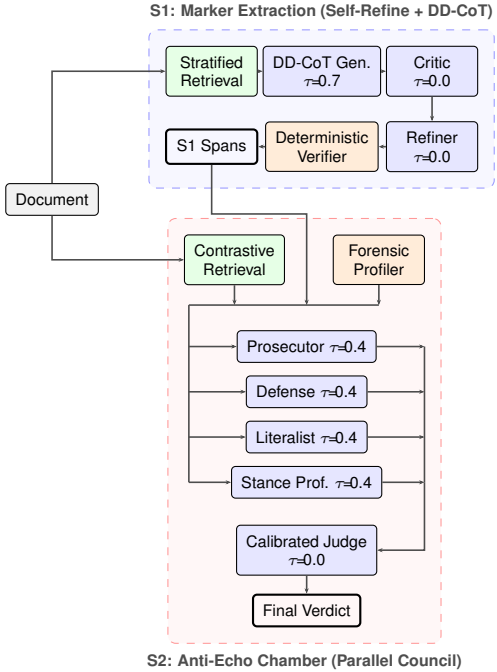
\begin{figure}[t]
    \centering
    \begin{tikzpicture}[
		>={Stealth[length=2pt]},
		inputbox/.style={rectangle, draw, rounded corners=1.5pt, minimum height=0.42cm,
				minimum width=1.2cm, align=center, font=\tiny\sffamily, fill=gray!10},
		ragbox/.style={rectangle, draw, rounded corners=1.5pt, minimum height=0.42cm,
				minimum width=1.2cm, align=center, font=\tiny\sffamily, fill=green!10},
		llmbox/.style={rectangle, draw, rounded corners=1.5pt, minimum height=0.42cm,
				minimum width=1.2cm, align=center, font=\tiny\sffamily, fill=blue!10},
		detbox/.style={rectangle, draw, rounded corners=1.5pt, minimum height=0.42cm,
				minimum width=1.2cm, align=center, font=\tiny\sffamily, fill=orange!15},
		outputbox/.style={rectangle, draw, rounded corners=1.5pt, minimum height=0.42cm,
				minimum width=1.2cm, align=center, font=\tiny\sffamily, thick},
		councilbox/.style={rectangle, draw, rounded corners=1.5pt, minimum height=0.42cm,
				minimum width=1.3cm, align=center, font=\tiny\sffamily, fill=blue!10},
		line/.style={-, semithick, black!70},
		arrow/.style={->, semithick, black!70},
		grouplab/.style={font=\bfseries\tiny\sffamily, text=black!70},
	]

		\node[inputbox] (input) at (0, -1.15) {Document};

		\node[ragbox] (s1rag) at (2.1, 0.55) {Stratified\\Retrieval};
		\node[llmbox] (gen) at (3.55, 0.55) {DD-CoT Gen.\\{\tiny $\tau$\!=\!0.7}};
		\node[llmbox] (critic) at (5.0, 0.55) {Critic\\{\tiny $\tau$\!=\!0.0}};
		\node[llmbox] (refiner) at (5.0, -0.45) {Refiner\\{\tiny $\tau$\!=\!0.0}};
		\node[detbox] (verifier) at (3.55, -0.45) {Deterministic\\Verifier};
		\node[outputbox] (s1out) at (2.1, -0.45) {S1 Spans};

		\draw[arrow] (s1rag) -- (gen);
		\draw[arrow] (gen) -- (critic);
		\draw[arrow] (critic) -- (refiner);
		\draw[arrow] (refiner) -- (verifier);
		\draw[arrow] (verifier) -- (s1out);

		\draw[arrow] (input.north) |- (s1rag.west);

		\begin{scope}[on background layer]
			\node[draw=blue!40, fill=blue!3, rounded corners=3pt, dashed,
			fit=(s1rag)(gen)(critic)(refiner)(verifier)(s1out),
			inner sep=6pt,
			label={[grouplab]above:{S1: Marker Extraction (Self-Refine + DD-CoT)}}] {};
		\end{scope}

		\node[ragbox] (s2rag) at (2.4, -1.95) {Contrastive\\Retrieval};
		\node[detbox] (forensic) at (4.3, -1.95) {Forensic\\Profiler};

		\node[councilbox] (pros) at (3.35, -3.1) {Prosecutor {\tiny $\tau$\!=\!0.4}};
		\node[councilbox] (defense) at (3.35, -3.7) {Defense {\tiny $\tau$\!=\!0.4}};
		\node[councilbox] (literal) at (3.35, -4.3) {Literalist {\tiny $\tau$\!=\!0.4}};
		\node[councilbox] (stance) at (3.35, -4.9) {Stance Prof. {\tiny $\tau$\!=\!0.4}};

		\begin{scope}[on background layer]
			\node[draw=blue!40, fill=blue!3, rounded corners=2pt, dashed,
			fit=(pros)(defense)(literal)(stance),
			inner sep=6pt,
			label={[grouplab]above:{Parallel Council}}] (pcgroup) {};
		\end{scope}

		\node[llmbox] (judge) at (3.35, -5.8) {Calibrated Judge\\{\tiny $\tau$\!=\!0.0}};
		\node[outputbox] (verdict) at (3.35, -6.6) {Final Verdict};

		\draw[arrow] (input.south) |- (s2rag.west);

		\coordinate (bus_l) at (1.8, -2.55);
		\coordinate (bus_r) at (4.3, -2.55);
		\draw[line] (bus_l) -- (bus_r);

		\draw[arrow] (s1out.south) -- ++(0,-0.65) -| (3.35, -2.55);
		\draw[arrow] (s2rag.south) -- (2.4, -2.55);
		\draw[arrow] (forensic.south) -- (4.3, -2.55);

		\coordinate (spine_bot) at (1.8, -4.9);
		\draw[line] (bus_l) -- (spine_bot);

		\foreach \j in {pros, defense, literal, stance} {
				\draw[arrow] (bus_l |- \j.west) -- (\j.west);
			}

		\coordinate (merge_top) at (4.9, -3.1);
		\coordinate (merge_bot) at (4.9, -4.9);
		\draw[line] (merge_top) -- (merge_bot);

		\foreach \j in {pros, defense, literal, stance} {
				\draw[arrow] (\j.east) -- (merge_top |- \j.east);
			}

		\draw[arrow] (merge_bot) |- (judge.east);

		\draw[arrow] (judge) -- (verdict);

		\begin{scope}[on background layer]
			\node[draw=red!40, fill=red!3, rounded corners=3pt, dashed,
			fit=(s2rag)(forensic)(pros)(stance)(judge)(verdict)(bus_l)(merge_top),
			inner sep=6pt,
			label={[grouplab]below:{S2: Anti-Echo Chamber (Parallel Council)}}] {};
		\end{scope}

	\end{tikzpicture}
    \vspace{-4pt}
    \caption{System architecture. \textbf{S1}: DD-CoT Self-Refine extracts markers; a deterministic verifier anchors them to character offsets. \textbf{S2}: Anti-Echo Chamber with contrastive retrieval, forensic profiling, a Parallel Council, and a calibrated Judge.}
    \label{fig:arch}
    \vspace{-8pt}
\end{figure}

We implement a two-stage agentic workflow: \textbf{S1} extracts
psycholinguistic marker spans, and \textbf{S2} predicts conspiracy endorsement
conditioned on the document and the extracted markers. The design separates (i)
LLM-mediated semantic decisions (what to extract / how to interpret stance)
from (ii) deterministic operations that require exactness (character offsets,
lightweight text statistics). Figure~\ref{fig:arch} summarizes the inference
flow.

\subsection{Contrastive Few-shot Retrieval}
\label{sec:contrastive-rag}
In-context few-shot examples are retrieved for the given document (no augmentation
is performed). A contrastive strategy is used to supply
discriminative precedents (including hard negatives for the Reporter Trap).
For S1, retrieval is additionally \emph{stratified}: we balance positive/negative documents and upweight underrepresented marker types (e.g., \textsc{Evidence}, \textsc{Victim}) so that prompts include instructive cases beyond the dominant \textsc{Actor}/\textsc{Action} patterns.
For S2, hard negatives are \texttt{non} documents that nevertheless contain S1
marker vocabulary; they match topic but oppose stance, forcing deliberation to
attend to attribution, hedging, and framing cues rather than topical overlap.
This targets Reporter Trap errors by construction.
Full retrieval specification and Figure~\ref{fig:rag} are
preserved in App.~\ref{app:rag-details}.

\subsection{S1: Marker Extraction via DD-CoT}

S1 produces a set of labeled spans by combining a self-refinement loop with a
deterministic span locator. The graph consumes the document and retrieved
few-shot precedents, generates candidate marker \emph{strings} with labels,
iteratively corrects them, then anchors each string to \texttt{startIndex} and
\texttt{endIndex} in the original text.

We explicitly decouple \textit{semantic identification} from \textit{span
	indexing}. LLMs can justify category assignments but are brittle at
character-accurate localization \cite{fu2024struggle}. We therefore (i) ask the
LLM to emit verbatim marker strings with labels, and (ii) compute offsets with
a deterministic locator that performs exact matching against the source text.
This avoids ``hallucinated spans'' and off-by-one indices while preserving the
LLM's interpretive signal \cite{ogasa-arase-2025-hallucinated}.


\paragraph{Dynamic Discriminative Chain-of-Thought (DD-CoT).} We introduce DD-CoT, which extends the adaptive reasoning of Dynamic CoT
\cite{ma2024dynamic} with an explicit \emph{discrimination step}. For each candidate span, the generator must state
(i) evidence for the chosen label and (ii) a short counter-argument against at
least one confusable label. This forces the model to commit to a decision
boundary in frequent confusions (e.g., \textsc{Actor} vs. \textsc{Victim},
\textsc{Action} vs.\ \textsc{Effect}) rather than producing post-hoc
rationales.


\paragraph{Agents.} The Self-Refine loop comprises four sequential nodes: (a)~a DD-CoT
\textbf{Generator} that proposes labeled marker strings; (b)~an
\textbf{Enhanced Critic} that checks verbatimness, boundaries, label
discrimination, and missing spans; (c)~a \textbf{Refiner} that applies
minimal edits; and (d)~a \textbf{Deterministic Verifier} that maps strings to
character offsets and deduplicates overlaps. The verifier's matching cascade is
detailed in Appendix~\ref{app:verifier}.

\paragraph{Self-Refine} follows the standard critique--revise pattern \cite{madaan2023selfrefine} but
operates over typed intermediate artifacts (candidate spans, critiques, and
edits), improving controllability and enabling deterministic verification.

\subsection{S2: Classification via Anti-Echo Chamber}

S2 targets a specific failure mode: \textbf{Reporter Trap} false positives, in
which topical discussion of conspiracies is conflated with endorsement (e.g.,
reporting, debunking, satire). Single-pass classifiers often over-commit early
and underweight stance cues \cite{wan-etal-2025-unveiling}. We therefore
structure S2 as a deterministic \emph{Forensic Profiler} that emits
stance-relevant warnings, a \emph{Parallel Council} that produces independent
pro/contra analyses, and a \emph{Calibrated Judge} that aggregates votes with
conservative confidence rules.

\paragraph{Forensic Profiler.}
Before LLM deliberation, a deterministic node computes lightweight linguistic
signals (e.g., attribution/reporting cues, shouting/affect, question-heavy
``JAQing'' (``just asking questions'') patterns) that are injected as
structured warnings (full  definitions are provided in
App.~\ref{app:forensic}).


\paragraph{Parallel Council Architecture.} The \textbf{Anti-Echo Chamber} enforces independent assessment by four personas
that receive identical inputs (document, S1 markers, retrieval context,
profiler warnings) and produce structured votes without seeing each other's
outputs: (1)~\textbf{Prosecutor} identifies evidence \emph{for} endorsement;
(2)~\textbf{Defense Attorney} presents evidence \emph{against} endorsement
(reporting/debunking/satire cues); (3)~\textbf{Literalist} independently checks
literal entailment and burden-of-proof on the source text; and
(4)~\textbf{Stance Profiler} analyzes stance cues (certainty, framing, group
dynamics) from the original document.
Each juror outputs a structured vote comprising a binary verdict, confidence
score, and textual evidence (App.~\ref{app:council-details}). This design reduces information leakage and ordering
effects typical of sequential debate.


\paragraph{Calibrated Judge.} The Judge aggregates votes using a weighted consensus score and applies
conservative adjudication rules to handle council splits and forensic warnings
(detailed in Appendix~\ref{app:council-details}). This ensures that the system
defaults to \texttt{non} when evidence is ambiguous or contradictory.

\section{Experimental Setup}

We follow the official SemEval train/dev splits without modification; we report
results on the provided dev set (\textit{100} documents) and the official test
set (\textit{938} documents). The primary baseline is a zero-shot GPT-5.2
classifier/extractor using a single prompt (no retrieval, no self-refinement,
no council). Our system uses the same base model but adds workflow structure
and contrastive retrieval.

\paragraph{Temperature Stratification.} To balance exploration and reproducibility (App.~\ref{app:exp-details}), we apply a differential temperature 
strategy. The S1 Generator uses $\tau=0.7$ to encourage diverse candidate
exploration for marker spans. Council Jurors use $\tau=0.4$ to integrate varied
rationales with consistent verdicts. Deterministic nodes  (Critic,
Refiner,  Judge) use $\tau=0.0$ to ensure consistent, reproducible auditing
and final adjudication. 


\paragraph{Contrastive sampling.} Retrieval draws in-context examples from a vector store with reranking, with an
emphasis on \emph{contrast} rather than similarity for S2. In particular, we
explicitly mine and retrieve hard negatives (\texttt{non} documents containing
S1 markers) to expose the stance boundary that drives Reporter Trap false
positives.

\paragraph{GEPA optimization.} Prompt templates are optimized with GEPA \cite{agrawal2025gepa}
(App.~\ref{sec:gepa-details}). We use a population of 20--30 prompts over
40--80 generations with tournament size 3, mutation rate 0.2, and GPT-5.2 for
semantic crossover and reflective mutation. S1 fitness is macro $F_2$
(\S\ref{sec:gepa-fitness}); S2 fitness uses \textbf{Gradient Consensus}
(vote-ratio scoring), where optimal prompts maximize the ratio of jurors
agreeing with the ground truth label. This continuous reward signal
distinguishes between weak (2--2) and strong (4--0) consensus, guiding the
optimizer toward robust prompts. Hyperparameters are summarized in
Table~\ref{tab:gepa-config}.
All agent nodes are instantiated via GPT-5.2 with schema-constrained generation via PydanticAI
\cite{pydanticai2024} and are orchestrated via LangGraph \cite{langgraph2024};
full implementation, preprocessing details, and reproducibility notes (multi-run validation for $\approx \pm 1.5\%$ F1 variance from MoE non-determinism) are in
App.~\ref{app:exp-details}.

\paragraph{Evaluation.}
For S1, we report the Macro Overlap F1 score, where an extracted span is considered a true positive if its character-level Intersection over Union (IoU) with a gold span is $\ge 0.5$. For S2, we report the macro-averaged F1-score; additional diagnostics (false positive rate on hard negatives) are reported where relevant.

\section{Results and Analysis}

\begin{table}[t!]
	\centering
	\small
	\setlength{\tabcolsep}{4pt}
	\begin{tabular}{@{}l l r r r@{}}
		\toprule
		\textbf{Task} & \textbf{Split}      & \textbf{Baseline F1} & \textbf{Agentic} & \textbf{$\Delta$} \\
		\midrule
		\multirow{2}{*}{S1}
		              & Dev (\textit{100})  & 0.12                 & \textbf{0.24}    & +100\%            \\
		              & Test (\textit{938}) & --                   & \textbf{0.21}    & --                \\
		\midrule
		\multirow{2}{*}{S2}
		              & Dev (\textit{100})  & 0.53                 & \textbf{0.79}    & +49\%             \\
		              & Test (\textit{938}) & --                   & \textbf{0.75}    & --                \\
		\bottomrule
	\end{tabular}
	\caption{Main results (macro F1). Baseline: zero-shot GPT-5.2. Document counts in parentheses.}
	\label{tab:main_results}
\end{table}

\begin{table*}[t!]
	\centering
	\small
	\begin{tabular}{p{2.2cm}p{6.8cm}lrrr}
		\toprule
& \textbf{Component / Change}         & \textbf{Metric} & \textbf{Baseline} & \textbf{Agentic} & \textbf{$\Delta$} \\
		\midrule
\multirow{3}{*}{\textit{Subtask 1:}} & \textbf{Self-Refine} (Audit loop)   & Macro F1        & 0.173             & 0.240            & +0.067            \\
 & \textbf{DD-CoT} (Perpetrator vs. Victim Discrimination)  & \textsc{Actor} F1        & 0.263             & 0.290            & +0.027            \\
\multirow{1}{*}{\textit{\newline Marker Extraction}} &\textbf{Contextual Retrieval} (Dynamic few-shot examples)       & Macro F1        & 0.243             & 0.240            & $-$0.003          \\
		\midrule
\multirow{2}{*}{\textit{Subtask 2:}} & \textbf{Parallel Council} (Debate)  & Recall          & 0.481             & 0.560            & +0.079            \\
& \textbf{Prosecutor} (Evidence for Conspiracy)        & F1 Score        & 0.680             & 0.795            & +0.115            \\
\multirow{1}{*}{\textit{Conspiracy}} & \textbf{Calibrated Judge} (Final Decision Logic) & Macro F1        & 0.638             & 0.681            & +0.043            \\
\multirow{1}{*}{\textit{Detection}}& \textbf{Contrastive Retrieval} (Suppression of False Positives)      & FP Rate         & 0.160             & 0.080            & $-$0.080          \\
		\bottomrule
	\end{tabular}
	\caption{\textbf{Ablation Summary (dev)} to isolate the impact of core architectural nodes. \textbf{DD-CoT} significantly improves \textsc{Actor} identification by disambiguating agency (detecting perpetrators vs. victims). \textbf{Parallel Council} and \textbf{Contrastive Retrieval} combine to suppress the ``Reporter Trap,'' where topical discussion of conspiracy is misclassified as endorsement.}
	\label{tab:combined_ablation}
\end{table*}

\subsection{Main Results}
\label{sec:main_results}

Table~\ref{tab:main_results} compares our workflow against the zero-shot baseline. The agentic pipeline doubles S1 performance 
(Dev F1 $0.12 \rightarrow 0.24$) by separating marker extraction from verification, while S2 gains ($0.53 \rightarrow 0.79$) 
are driven by the Council's ability to resolve stance ambiguity. Specifically, DD-CoT improves \textsc{Actor} identification (+2.7 F1) 
by disambiguating agency in passive structures, and the Council-Judge architecture significantly increases recall (+16.4\%) while suppressing 
false positives via contrastive retrieval. In the official evaluation, our system ranked 3rd on the S1 development set and 10th on the test set (within $0.05~F1$ of the top system); for S2 it placed 13th and 24th, respectively.

\paragraph{Open-Weights Models.} To test portability, we deploy simplified agentic schemas on Qwen-3-8B-Instruct; core reasoning transfers effectively despite reduced schema fidelity (App.~\ref{app:open-weights}).


\paragraph{Ablation Analysis.}

Table~\ref{tab:combined_ablation} highlights three takeaways. \textbf{Iteration
	matters:} Self-Refine yields the largest S1 gain (+6.7\% F1) by correcting
boundaries. \textbf{Discrimination improves agency:} DD-CoT overcomes
subject-position bias for a +2.7\% \textsc{Actor} F1 gain. \textbf{Architecture synergy is
	critical:} In S2, contrastive retrieval halves the false-positive rate,
while the Parallel Council improves recall (+16\%). Leave-One-Out analysis
(Figure~\ref{fig:s2_juror_ablation}, Appendix~\ref{app:s2-details}) confirms
this synergy: removing the \textbf{Prosecutor} drops F1 by $\sim$11\%,
while removing skeptical jurors harms precision. Finally, \textbf{Contextual Retrieval}
boosted generalization on the Test set significantly (+10.5\% F1) despite
negligible impact on Dev.

\paragraph{Qualitative Analysis.} As detailed in App.~\ref{app:qualitative-examples}, the agentic pipeline succeeds by
disentangling agency and mitigating the ``Reporter Trap.'' Unlike baselines that conflate grammatical
subjects with semantic agents, DD-CoT correctly identifies \textsc{Actor}s via semantic roles.
For S2, contrastive retrieval and defensive parsing of attribution verbs (\textit{claimed}, \textit{reported})
prevent false endorsement on neutral reporting. However, high-context irony remains challenging.

\paragraph{Sensitivity and Robustness.} Temperature stratification prevents Council mode collapse (redundant arguments at $\tau\!=\!0.0$) and preserves boundary enforcement (erosion at $\tau\!>\!0.7$ for Judge/Critic). The Judge is resistant to relaxed thresholds, though lowering the override threshold slightly increases false positives.

\paragraph{Computational Cost and Latency.}

We profile baseline and full-pipeline overhead on a 20-document dev
sample (Table~\ref{tab:latency}). The S1 Self-Refine loop triples latency ($10.5$s $\to$ $30.2$s/doc)
and increases token use $2.9\times$. For S2, the Parallel Council adds $6.4\times$ latency ($4.6$s $\to$ $29.1$s)
and $7.6\times$ tokens relative to a single-agent baseline. Crucially, by fixing the number of LLM calls,
latency remains \emph{constant per document}, avoiding the unbounded scaling of recursive generation schemes.

\begin{table}[t!]
	\centering
	\small
	\begin{tabular}{p{0.1cm}p{3.7cm}p{0.8cm}p{0.6cm}r}
		\toprule
		\textbf{} & \textbf{Configuration} & \textbf{Latency} & \textbf{Tokens} & \textbf{Calls} \\
		\midrule
		\multirow{2}{*}{S1} & Baseline (Generator only) & 10.5s & 2{,}549 & 1.0 \\
		                    & Full Graph (Self-Refine)   & 30.2s & 7{,}986 & 2.5 \\
		\midrule
		\multirow{2}{*}{S2} & Baseline (Single Agent)       & 4.6s  & 1{,}665 & 1.0 \\
		                    & Full Graph (Council + Judge)   & 29.1s & 12{,}627 & 5.0 \\
		\bottomrule
	\end{tabular}
	\caption{Per-document average latency and token usage profiled on 20 dev documents. All calls are LLM API invocations; baseline is a single zero-shot prompt.}
	\label{tab:latency}
\end{table}

\section{Conclusion}

We demonstrate that \textbf{workflow structure} can substitute for model scaling in psycholinguistic NLP: DD-CoT with deterministic anchoring doubles S1 performance, while an adversarial Parallel Council suppresses ``Reporter Trap'' false positives for S2. More broadly, structured multi-agent deliberation recovers performance lost to task complexity without larger models or additional training data, though its effectiveness depends critically on persona diversity and retrieval quality. We note that conspiracy detection tools carry inherent dual-use risks and recommend human-in-the-loop oversight for deployment.

\section*{Limitations}

Our pipeline operates purely via prompting and agentic orchestration; we did not fine-tune any model on the task data, which could improve both marker boundary precision and stance classification through task-specific supervision.
We also did not incorporate discourse-level context such as thread structure, parent comments, or user posting history, which could help disambiguate irony, sarcasm, and implicit stance, our primary failure modes.
Additionally, we did not perform systematic human evaluation of extracted markers; such analysis could quantify interpretability gains beyond automated overlap metrics.
While we experimented with a single open-weights model (Qwen-3-8B), we did not explore ensembles of heterogeneous models or distillation strategies to reduce dependence on a proprietary backbone while preserving agentic reasoning.
Finally, we did not investigate data augmentation (e.g., synthetic hard negatives or paraphrase-based span perturbations) to improve robustness to the lexical and stylistic variation observed between dev and test sets.

\typeout{MAIN_BODY_LAST_PAGE=\thepage}
\bibliography{custom}

\appendix
\newpage
\typeout{APPENDIX_START_PAGE=\thepage}

\section{Task and Background Details}
\label{app:task-details}

\paragraph{Marker taxonomy (S1).}
S1 annotates spans belonging to: \textsc{Actor} (agent), \textsc{Action} (what
the actor does), \textsc{Effect} (consequences), \textsc{Victim} (who is
harmed), \textsc{Evidence} (claims or proof used to support the theory).

\section{Deterministic Verifier Details}
\label{app:verifier}
The Deterministic Verifier is a non-LLM post-processing node that serves as
the \textit{structural locator}, anchoring LLM-generated text strings to
character-precise offsets through a five-tier matching cascade:
\begin{enumerate}
	\item[(i)] \textbf{Exact match:} Byte-for-byte substring search supporting nth-occurrence disambiguation.
	\item[(ii)] \textbf{Case-insensitive:} Unicode-safe lowered comparison with original-position projection.
	\item[(iii)] \textbf{Normalized:} Smart-quote straightening, whitespace collapse, and lowering with index remapping to recover original character offsets.
	\item[(iv)] \textbf{Fuzzy (Levenshtein):} Approximate matching with maximum edit distance $\leq 15\%$ of snippet length (minimum~1), activated only for spans $>$4 characters to avoid spurious short matches.
	\item[(v)] \textbf{SequenceMatcher alignment:} LCS-based last-resort recovery requiring $\geq 60\%$ character coverage and compactness $\leq 1.5\times$ snippet length, with word-boundary snapping.
\end{enumerate}
Each tier is attempted in order; the first successful match is accepted.
Additionally, the Verifier implements aggressive cross-label deduplication to
eliminate overlapping or duplicate spans. This graduated recovery strategy
ensures that all submitted spans are structurally valid and anchored to the
source text, even when upstream LLM paraphrases or reformulations introduce
minor textual divergence from the original document.

\section{Forensic Profiler Metrics}
\label{app:forensic}
\paragraph{Forensic Profiler.} A deterministic profiling node computes linguistic signatures \textit{before}
any LLM deliberation, providing objective textual evidence to ground council
reasoning. Six metrics are extracted, each normalized by total word count
$|W|$:
\begin{enumerate}
	\item \textbf{Attribution Density:} $\text{AD} = |\{w \in W : w \in
		      V_{\text{attr}}\}| \,/\, |W|$, where $V_{\text{attr}}$ includes
	      distancing verbs (\textit{said}, \textit{claimed}, \textit{according
		      to}, \textit{reported}, \textit{sources}). Texts with AD $> 3.5\%$
	      receive an explicit \textsc{Reporter\_Warning} flag, signaling likely
	      journalistic framing rather than endorsement.
	\item \textbf{Shouting Score:} $\text{SS} = |\{w \in W : w =
		      \texttt{UPPER}(w) \land |w| > 1\}| \,/\, |W|$. Scores exceeding $10\%$
	      trigger an \textsc{Emotional\_Intensity} flag, as ALL-CAPS usage
	      correlates with conspiratorial conviction.
	\item \textbf{JAQing Detection:} A boolean flag activated when question
	      density $> 0.35$ (questions per sentence) \textit{and} hedging ratio
	      $> 5\%$ (terms like \textit{maybe}, \textit{perhaps}, \textit{just
		      asking}), identifying the ``Just Asking Questions'' rhetorical
	      manipulation pattern.
	\item \textbf{Agency Gap:} Passive voice proxy computed as $|\{w \in W : w
		      \in \{\textit{been}, \textit{being}, \textit{was}, \textit{were},
		      \textit{by}\}\}| \,/\, |W|$. Values $> 6\%$ suggest hidden agency
	      attribution, a hallmark of conspiratorial framing where actors are
	      deliberately obscured.
	\item \textbf{Epistemic Intensity:} Frequency of truth-claiming terms
	      (\textit{proof}, \textit{truth}, \textit{exposed}, \textit{revealed},
	      \textit{undeniable}) normalized by $|W|$, capturing the degree of
	      conspiratorial conviction expressed through epistemic certainty.
	\item \textbf{Question Density:} Number of question marks per sentence,
	      used as a component of JAQing detection and independently injected
	      into the Judge's case file for calibration.
\end{enumerate}
These metrics are injected into the Calibrated Judge's case file as structured
contextual warnings (e.g., \texttt{REPORTER\_WARNING: Attribution
	Density=4.2\%}). Council jurors receive forensic context indirectly through
enhanced marker summaries that include active warnings (e.g., high attribution
or JAQing patterns detected by the Forensic Profiler node), providing
deterministic anchors that constrain LLM reasoning.

\section{Contrastive Few-shot Retrieval Details}
\label{app:rag-details}
Both subtasks employ dynamic few-shot retrieval from ChromaDB \cite{chromadb2023} vector collections.
We retrieve \textbf{in-context few-shot examples} relevant to the given document to guide the model's decision (we do not perform augmentation). Inspired by Contrastive Chain-of-Thought prompting \cite{chia2023contrastive}, which enhances reasoning by supplying both valid and invalid demonstrations, we implement a contrastive a retrieval strategy that prioritizes discriminative examples over merely similar ones. Both S1 and S2 employ contrastive retrieval mechanisms, though optimized for different discriminative objectives.

\paragraph{S1: Stratified Contrastive Sampling.}
For marker extraction, we implement a dual-axis contrastive strategy. First, we
retrieve \textbf{balanced positive and negative examples} (documents labeled as
conspiracy and non-conspiracy) to teach the model that psycholinguistic markers
can appear in \textit{both} contexts (e.g., news articles may contain
\textsc{Actor} mentions without endorsing conspiracy). Second, within these
retrieved examples, we apply \textbf{marker-type stratification}, allocating
60\% retrieval weight to underrepresented categories (\textsc{Evidence} and
\textsc{Victim}), ensuring sufficient exposure to rare marker types. Finally,
all candidates undergo \textbf{cross-encoder reranking}
(BAAI/bge-reranker-v2-m3) \cite{bge_m3} to prioritize examples with similar
discourse structure over mere lexical overlap. This three-stage pipeline
addresses both label imbalance and annotation granularity mismatches.

\begin{figure}[t]
	\centering
	\small
	\resizebox{\linewidth}{!}{
		\begin{tikzpicture}[
			node distance=0.5cm and 0.4cm,
			>={Stealth[length=3pt]},
			box/.style={rectangle, draw, rounded corners=2pt, minimum height=0.6cm,
					align=center, font=\scriptsize\sffamily, inner sep=2pt},
			sbox/.style={box, fill=blue!10, minimum width=2.2cm},
			nbox/.style={box, fill=red!10, minimum width=2.2cm},
			gbox/.style={box, fill=green!10, minimum width=2.8cm},
			arrow/.style={->, thick, black!70},
			]

			\node[box, fill=gray!15, minimum width=2.5cm] (query) {Input Document};
			\node[box, fill=yellow!15, below=0.5cm of query, minimum width=2.5cm] (chroma) {ChromaDB Embeddings};
			\draw[arrow] (query) -- node[right, font=\tiny\sffamily] {embed} (chroma);

			\node[sbox, below left=0.6cm and 0.3cm of chroma] (s1ret) {Balanced + Type\\Stratification (60\%)};
			\node[nbox, below right=0.6cm and 0.3cm of chroma] (s2ret) {Hard Negatives\\(Non-CT + Markers)};
			\draw[arrow] (chroma.south) -- ++(-0.1,-0.2) -| (s1ret.north);
			\draw[arrow] (chroma.south) -- ++(0.1,-0.2) -| (s2ret.north);

			\node[gbox, below=0.5cm of s1ret] (rerank1) {Cross-Encoder\\Reranking (3×)};
			\node[gbox, below=0.5cm of s2ret] (rerank2) {Reranking (4×)\\+ Filtering};
			\draw[arrow] (s1ret) -- (rerank1);
			\draw[arrow] (s2ret) -- (rerank2);

			\node[box, fill=blue!5, below=0.5cm of rerank1, minimum width=2.2cm] (out1) {S1 Few-Shots};
			\node[box, fill=red!5, below=0.5cm of rerank2, minimum width=2.2cm] (out2) {S2 Precedents};
			\draw[arrow] (rerank1) -- (out1);
			\draw[arrow] (rerank2) -- (out2);

		\end{tikzpicture}
	}
	\caption{Contrastive few-shot retrieval architecture.}
	\label{fig:rag}
\end{figure}
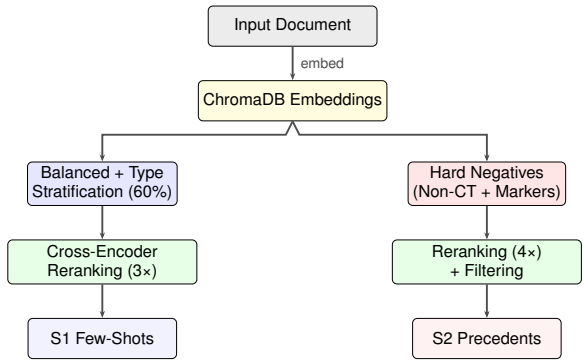

The overall contrastive retrieval strategy is illustrated in
Figure~\ref{fig:rag}.

\paragraph{S2: Hard Negative Mining \cite{karpukhin-etal-2020-dense}.}
For conspiracy detection, we implement a \textbf{pure contrastive strategy} via hard negative mining. Standard similarity-based retrieval retrieves similar-looking documents, causing the model to conflate \textit{topical similarity} with \textit{stance endorsement}, a failure mode we term the Reporter Trap. To explicitly teach the boundary, we retrieve documents labeled ``non-conspiracy'' \textit{that contain S1 markers}. These are \textit{hard} negatives because they share conspiracy-related vocabulary (actors, actions, evidence mentions) but differ in stance (reporting, debunking, or mocking). By forcing the model to compare structurally similar examples with opposite labels, we compel it to attend to \textbf{stance cues} such as attribution verbs (``claims that'', ``alleges''), hedging markers (``supposedly'', ``according to''), and framing signals (``debunked'', ``baseless''), rather than mere topic keywords. Retrieved precedents follow case-law templates (e.g., ``Acquitted because the text attributes claims without endorsement'') that provide structured reasoning patterns. Candidates undergo the same cross-encoder reranking with an elevated
overretrieve factor ($4\times$ vs.\ $3\times$ for S1). The higher factor
reflects the scarcity of high-quality hard negatives: because truly
contrastive examples (non-conspiratorial texts that nonetheless contain
conspiracy-related vocabulary) are rare in the training distribution
($<$20\% of documents), casting a wider retrieval net is necessary to ensure
the final prompt contains sufficiently discriminative pairs. Label-balanced
filtering maintains equal representation of hard negatives and true positives
in the final prompt context.

\section{Experimental Details}
\label{app:exp-details}
\paragraph{Dataset.}
All experiments use the official training/development splits without
modification. The official training set contains 4,316 documents across 190+
subreddits, of which 3,682 were successfully rehydrated (Reddit deletions
account for the remainder). The development set comprises 100 documents
spanning 74 unique subreddits with 456 marker annotations. In the development
set, the marker type distribution shows \textsc{Actor} (29.8\%) and
\textsc{Action} (22.8\%) dominating ($\sim$53\% combined), while
\textsc{Evidence} (16.0\%), \textsc{Victim} (15.8\%), and \textsc{Effect}
(15.6\%) are more balanced. The training set exhibits more severe imbalance
with \textsc{Actor} and \textsc{Action} comprising $\sim$70\% combined. This
skew motivates our stratified sampling strategy in the few-shot retrieval
component, allocating 60\% retrieval weight to underrepresented categories. For
S2, the development labels are distributed as: No (50.0\%), Yes (27.0\%), and
Can't Tell (23.0\%). The \textit{hard negative} subset (texts discussing
conspiracies without endorsing them) comprises $<$20\% of the training data,
necessitating explicit hard negative mining. A detailed exploratory analysis is
provided in Appendix~\ref{sec:eda}.

\paragraph{Data Preprocessing.}
Since individual documents may have multiple annotators, we apply
\textbf{majority-vote consensus} at both document and span level. For document
labels, the most frequent annotation is selected and exact ties are discarded.
For spans, overlapping annotations of the same marker type are clustered by
character overlap; clusters reaching the majority threshold (over half of
annotators) produce a single representative span (the longest in the cluster),
while sub-threshold clusters are dropped. This yields deterministic,
high-agreement annotations suitable for both training and few-shot retrieval.
After consensus, we remove near-duplicate documents via locality-sensitive
hashing (LSH, 8 bands), reducing the training set from 3,682 rehydrated
documents to 3,271 unique instances. \textit{Can't Tell} documents (607 in
training, $\sim$18.6\%) are handled asymmetrically: they are \textbf{retained
	for S1} (marker extraction can still learn from ambiguous texts containing
valid spans) but \textbf{excluded from S2} (conspiracy detection requires a
binary ground truth). Additionally, documents with no annotated spans and no
annotator disagreement are included in the S1 training corpus with 15\%
probability, serving as negative calibration examples that teach the Generator
to produce empty extractions for non-conspiratorial text. For S2 corpus
curation, a subtype-stratified sampling strategy selects documents across six
rhetorical subtypes (hard negatives, mundane negatives, debunking negatives,
evangelist conspiracy, insider conspiracy, and general conspiracy) to ensure
balanced exposure during prompt optimization, with hard negatives defined
broadly to include both non-conspiratorial texts containing markers
\textit{and} texts matching debunking-vocabulary cues.

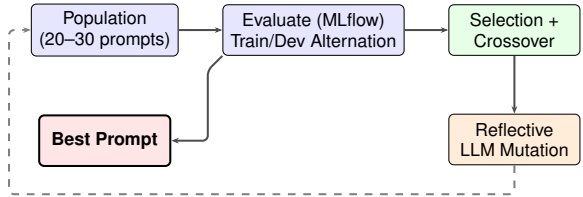
\begin{figure}[t]
	\centering
	\resizebox{\linewidth}{!}{
		\small
		\begin{tikzpicture}[
			node distance=0.6cm and 0.5cm,
			>={Stealth[length=3pt]},
			process/.style={rectangle, draw, rounded corners=2pt, minimum height=0.7cm,
					minimum width=1.8cm, align=center, font=\scriptsize\sffamily, fill=blue!10},
			decision/.style={rectangle, draw, rounded corners=2pt, minimum height=0.7cm,
					minimum width=1.8cm, align=center, font=\scriptsize\sffamily, fill=green!10},
			mutation/.style={rectangle, draw, rounded corners=2pt, minimum height=0.7cm,
					minimum width=1.8cm, align=center, font=\scriptsize\sffamily, fill=orange!15},
			output/.style={rectangle, draw, rounded corners=2pt, minimum height=0.7cm,
					minimum width=1.8cm, align=center, font=\bfseries\scriptsize\sffamily,
					fill=red!10, thick},
			arrow/.style={->, thick, black!70, rounded corners=2pt},
			dasharrow/.style={->, thick, dashed, black!50, rounded corners=2pt},
			]

			\node[process] (pop) {Population\\(20--30 prompts)};
			\node[process, right=0.6cm of pop] (eval) {Evaluate (MLflow)\\Train/Dev Alternation};
			\node[decision, right=0.6cm of eval] (select) {Selection +\\Crossover};
			\node[mutation, below=0.8cm of select] (mutate) {Reflective\\LLM Mutation};
			\node[output, below=0.8cm of pop] (best) {Best Prompt};

			\draw[arrow] (pop) -- (eval);
			\draw[arrow] (eval) -- (select);
			\draw[arrow] (select) -- (mutate);

			\draw[dasharrow] (mutate.south) -- ++(0,-0.4)
			-| ([xshift=-0.4cm]best.west)
			|- (pop.west);
			\draw[arrow] (eval.south west) -- ++(-0.2,-0.2) |- (best.east);

		\end{tikzpicture}
	}
	\caption{GEPA prompt optimization workflow.}
	\label{fig:gepa}
\end{figure}

\paragraph{Pipeline Components.} All final experiments use OpenAI \textbf{GPT-5.2} accessed via Pydantic-AI
\cite{pydanticai2024} for schema-constrained generation. Stateful agent
workflows are implemented as directed acyclic graphs using \textbf{LangGraph}
\cite{langgraph2024}, where each node maintains typed state with explicit field
annotations enabling deterministic transitions. The few-shot retrieval
component uses \textbf{ChromaDB} \cite{chromadb2023} with OpenAI
text-embedding-3-small embeddings (1536 dimensions) and \textbf{Maximal
	Marginal Relevance (MMR)} reranking \cite{carbonell1998mmr} using the
\textbf{BAAI/bge-reranker-v2-m3} cross-encoder. MMR balances relevance against
diversity via:

\begin{equation}
	\resizebox{0.9\linewidth}{!}{$
			\text{MMR} =
			\arg\max_{d_i \in R \setminus S}\big[\lambda \cdot \text{Rel}(d_i, q) -
				(1-\lambda) \cdot \max_{d_j \in S}\text{Sim}(d_i, d_j)\big]
		$}
\end{equation}

where $R$ is the candidate set, $S$ the already-selected documents, and
$\lambda=0.7$ biases toward relevance while preventing near-duplicate
few-shots. Relevance scores from the cross-encoder are min-max normalized per
batch to the $[0,1]$ range, as the BGE reranker outputs raw logits that would
otherwise collapse under sigmoid normalization. S1 retrieval over-retrieves
$3\times$ candidates before reranking, while S2 uses $4\times$ to ensure
higher-quality hard negatives. All LLM calls execute asynchronously with
exponential backoff retry logic (base 2s, max 5 retries). We employ
differential temperature settings: $\tau=0.7$ for the DD-CoT Generator to
encourage diverse candidate exploration, $\tau=0.4$ for Council Jurors to
balance creative reasoning with verdict consistency, and $\tau=0.0$ for the
Critic, Refiner, and Judge to enforce deterministic, reproducible auditing.
This stratification reflects each agent's functional role: generative nodes
benefit from sampling diversity to avoid mode collapse over marker types, while
evaluative nodes require strict adherence to textual evidence. For prompt
optimization, we utilize \textbf{GEPA} \cite{agrawal2025gepa} integrated with
MLflow \cite{mlflow2024}, using a passthrough injection pattern to tunnel gold
labels through the prediction wrapper for custom scoring. We conduct
optimization runs targeting S1 and S2 system prompts with population sizes of
20--30 candidates and 40--80 trials per run, alternating between training and
development splits to ensure generalization. Final prompts achieved +4.2\%
absolute $F_1$ improvement over hand-crafted baselines. The GEPA optimization
workflow is illustrated in Figure~\ref{fig:gepa}.

\section{Detailed Ablation Studies}
\label{app:ablation-full}
This section provides comprehensive tables and detailed narratives supporting the ablation studies discussed previously.

\subsection{S1 Agent Ablation}
Table~\ref{tab:s1_agents_detailed} isolates the contribution of each agent in
the S1 Self-Refine loop. The \textbf{Generator} provides the initial recall
base. The \textbf{Critic} significantly improves precision by filtering
hallucinated spans, while the \textbf{Refiner} optimizes boundaries
(startIndex/endIndex), providing a smaller but critical boost to exact-match
F1.

\begin{table}[h]
	\centering
	\small
	\resizebox{\linewidth}{!}{
		\begin{tabular}{@{}lrrr@{}}
			\toprule
			\textbf{Configuration} & \textbf{Precision} & \textbf{Recall} & \textbf{Macro F1} \\
			\midrule
			Generator Only (Base)  & 0.145              & 0.215           & 0.173             \\
			+ Enhanced Critic      & 0.198              & 0.225           & 0.211             \\
			+ Refiner (Full S1)    & \textbf{0.221}     & \textbf{0.262}  & \textbf{0.240}    \\
			\bottomrule
		\end{tabular}
	}
	\caption{S1 Agent Ablation (Dev Set). Breakdown of contributions from the Critic and Refiner agents.}
	\label{tab:s1_agents_detailed}
\end{table}

\subsection{Juror Ablation Study (S2)}
\label{app:s2-details}
We modified the \texttt{run\_s2\_parallel\_council} function to support dynamic
juror selection, allowing us to test the impact of specific personas
(Prosecutor, Defense, Literalist, Profiler) on the final verdict using a
``Leave-One-Out'' (LOO) methodology.

\paragraph{Scientific Interpretation.} The Leave-One-Out analysis demonstrates the synergistic robustness of the
Parallel Council architecture (Baseline F1: 0.795):
\begin{enumerate}
	\item \textbf{Primary Signal Driver (Prosecutor):} Removing the Prosecutor results in an $\sim$11\% drop in F1, as the system loses its primary mechanism for identifying latent conspiracy markers.
	\item \textbf{Adversarial Balance (Defense/Literalist):} The removal of these skeptical roles degrades Precision and Negative Predictive Value. False Positives increase as the model lacks the necessary adversarial checks to differentiate between reporting and endorsement.
	\item \textbf{The Importance of Profiling:} The $\sim$5\% drop when removing the Profiler indicates the value of contextual subreddit priors and linguistic intensity metrics in adjudicating borderline cases.
\end{enumerate}

\begin{figure}[h]
	\centering
	\begin{tikzpicture}
		\begin{axis}[
			ybar,
			bar width=6pt,
			width=\linewidth,
			height=5cm,
			enlarge x limits=0.15,
			legend style={at={(0.5,1.15)}, anchor=north, legend columns=-1, font=\scriptsize},
			symbolic x coords={Full Council, No Prosecutor, No Defense, No Literalist, No Profiler},
			xtick=data,
			x tick label style={rotate=30, anchor=east, font=\scriptsize},
			ymin=0.5, ymax=1.0,
			ylabel={Score},
			ymajorgrids=true,
			grid style=dashed
		]
		\addplot coordinates {(Full Council, 0.795) (No Prosecutor, 0.680) (No Defense, 0.620) (No Literalist, 0.695) (No Profiler, 0.740)};
		\addplot coordinates {(Full Council, 0.880) (No Prosecutor, 0.773) (No Defense, 0.714) (No Literalist, 0.818) (No Profiler, 0.870)};
		\addplot coordinates {(Full Council, 0.815) (No Prosecutor, 0.630) (No Defense, 0.556) (No Literalist, 0.667) (No Profiler, 0.741)};
		\legend{F1, Precision, Recall}
		\end{axis}
	\end{tikzpicture}
	\vspace{-5pt}
	\caption{Juror Ablation Study (LOO). Removing any single persona significantly degrades performance, validating the necessity of the full four-juror council.}
	\label{fig:s2_juror_ablation}
\end{figure}
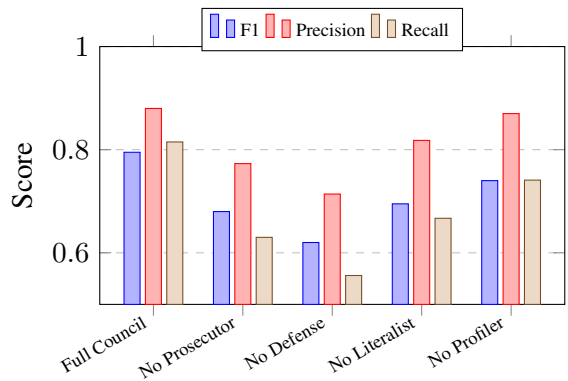

\begin{table}[h]
	\centering
	\small
	\resizebox{\linewidth}{!}{
	\begin{tabular}{@{}lccc@{}}
		\toprule
		\textbf{Metric} & \textbf{Standard CoT (Base)} & \textbf{DD-CoT (New)} & \textbf{$\Delta$} \\
		\midrule
		\textsc{Actor} F1        & 26.3\%                       & 29.0\%                & +2.7 pts          \\
		\textsc{Victim} F1       & 23.3\%                       & 22.0\%                & $-$1.3 pts        \\
		\bottomrule
	\end{tabular}
	}
	\caption{Impact of DD-CoT on Agency Disentanglement (dev). DD-CoT significantly improves \textsc{Actor} detection by resolving subject-position bias, with a minor redistribution of Victim scores.}
	\label{tab:cot_ablation_detailed}
\end{table}

\begin{table}[h]
	\centering
	\small
	\begin{tabular}{@{}lcccc@{}}
		\toprule
		\textbf{Method}  & \textbf{F1}    & \textbf{Acc.}  & \textbf{Deadlock} & \textbf{FP Conf.} \\
		\midrule
		Majority Vote    & 0.638          & 0.779          & 66.7\%            & 0.890             \\
		Calibrated Judge & \textbf{0.681} & \textbf{0.805} & \textbf{100.0\%}  & \textbf{0.865}    \\
		\bottomrule
	\end{tabular}
	\caption{Ablation: Judge vs.\ Majority Vote (original).}
	\label{tab:ablation_judge}
\end{table}

\begin{table}[h]
	\centering
	\small
	\begin{tabular}{@{}lccc@{}}
		\toprule
		\textbf{Configuration}      & \textbf{Macro F1} & \textbf{Precision} & \textbf{Recall} \\
		\midrule
		No Retrieval                & \textbf{0.329}    & 0.322              & \textbf{0.419}  \\
		Standard Retrieval (Naive)  & 0.296             & 0.292              & 0.376           \\
		Stratified Retrieval (Ours) & 0.311             & \textbf{0.313}     & 0.392           \\
		\bottomrule
	\end{tabular}
	\caption{Few-shot retrieval strategy comparison on dev set (S1) (original).}
	\label{tab:s1_rag_ablation}
\end{table}

\begin{table}[h]
	\centering
	\small
	\begin{tabular}{@{}lcc@{}}
		\toprule
		\textbf{Configuration} & \textbf{False Positive Rate} & \textbf{Reduction} \\
		\midrule
		Standard Retrieval     & 0.160                        & --                 \\
		Contrastive Retrieval  & \textbf{0.080}               & \textbf{50\%}      \\
		\bottomrule
	\end{tabular}
	\caption{S2 Retrieval Ablation: Impact of Contrastive Retrieval on False Positive Rate (original).}
	\label{tab:rag_ablation}
\end{table}

\section{Extended Methodological Narratives}
\label{app:extended-narratives}
The following section provides a more detailed elaboration of our methodological approaches and results.

\paragraph{Extended Main Results Narrative.}
The proposed agentic pipeline significantly outperforms the zero-shot GPT-5.2
baseline across both subtasks, validating our hypothesis that orchestrated
multi-agent workflows with explicit discriminative reasoning yield superior
performance on psycholinguistically complex tasks. Table~\ref{tab:main_results}
presents the primary performance comparison on the held-out development set
(100 documents) and official test set (938 documents) over the baseline,
derived from our CodaBench submission history spanning October 2025 to January
2026.

\textbf{S1: Marker Extraction}
performance \textbf{doubled} (F1: from 0.12 to 0.24 on dev), demonstrating that simple zero-shot prompting fails to capture the
complexity of psycholinguistic span extraction. Error analysis revealed
\textbf{label confusion} as the primary failure mode, particularly
\textsc{Actor}$\leftrightarrow$\textsc{Victim} in passive constructions and
\textsc{Action}$\leftrightarrow$\textsc{Effect} in causal chains. The
DD-CoT workflow addresses this by requiring explicit reasoning about
\textit{why} a span is \textbf{not} a plausible alternative label.

\textbf{S2: Conspiracy Detection}
F1 improved from 0.53 to 0.79 (+49\% relative). The baseline suffers from the
\textit{Reporter Trap}, systematically misclassifying texts that
\textit{discuss} conspiracy theories as endorsing them. Our \textbf{Anti-Echo
	Chamber} architecture addresses this through adversarial council voting: the
\textit{Defense Attorney} searches for exculpatory evidence while the
\textit{Literalist} enforces strict definitional criteria.

\paragraph{Test Set Generalization.}
Both S1, S2 slightly degrade on the larger test set (S1: -12.5\%, S2: -5\%),
consistent with distribution shift and the inherent difficulty of span
extraction on unseen text.

\paragraph{Extended Conclusion.}
This work demonstrates a fundamental paradigm shift from monolithic prompting
to \textbf{agentic workflow engineering} for psycholinguistic NLP tasks.
Complex discriminations such as distinguishing \textsc{Actor} from
\textsc{Victim} or \textit{topical discussion} from \textit{stance endorsement}
cannot be resolved by a single ``perfect prompt.'' Instead, they require a
\textbf{chain of responsibility} where specialized agents execute complementary
functions: generation, critique, refinement, and verification. For Subtask 1,
we addressed the ``Hallucinated Span'' problem by coupling a \textbf{Semantic
	Reasoner} (DD-CoT) with a \textbf{Deterministic Locator}, achieving a
\textbf{doubling of F1 performance} (from 0.12 to 0.24) while eliminating
character-level indexing errors. The explicit discriminative reasoning
mechanism,requiring the model to articulate ``Why NOT'' alternative labels
proved essential for agency detection, yielding a +2.7 point gain in \textsc{Actor} F1
(Table~\ref{tab:combined_ablation}). For Subtask 2, the \textbf{Anti-Echo Chamber}
architecture (Parallel Council + Calibrated Judge) successfully disentangled
conspiracy \textit{topics} from conspiratorial \textit{stance}, overcoming the
Reporter Trap that plagued single-agent classifiers. Critically, the Calibrated
Judge achieved \textbf{100\% accuracy on deadlocks}
(Table~\ref{tab:ablation_judge}), demonstrating that AI can resolve its own
ambiguity when provided with structured debate transcripts rather than mere
vote counts. Our \textbf{Contrastive few-shot retrieval} strategy, hard
negative mining combined with stratified sampling, reduced the False Positive
Rate by 50\% (from 0.160 to 0.080), validating that retrieval strategy design
is as critical as retrieval presence itself.

While the system improves both extraction fidelity and stance discrimination,
we find that pragmatic phenomena such as high-context irony remain challenging
without external user/discourse context (Appendix~\ref{app:qualitative-examples}).

\section{Exploratory Data Analysis}
\label{sec:eda}

\begin{figure}[h!]
	\centering
	\includegraphics[width=\linewidth]{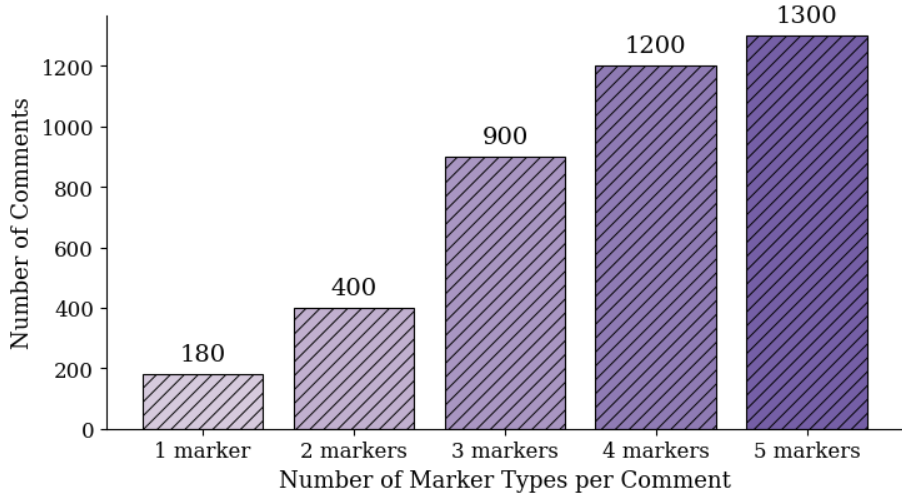}
	\caption{Number of marker types in the dataset.}
	\label{fig:num-of-markers}
\end{figure}

\paragraph{Annotation Coverage}
As mentioned in the official website of the task, there are more than 4,100
unique Reddit comments, including 4,800 annotations in total. Most comments
($\sim$3,500), have only one annotation, 550 have two, and 50 have more.
Regarding marker density, around 4,000 comments have at least one
psycholinguistic marker annotation. The exact distribution of marker category
coverage in comments is demonstrated in Figure \ref{fig:num-of-markers}.

\begin{figure}[h!]
	\centering
	\includegraphics[width=\linewidth]{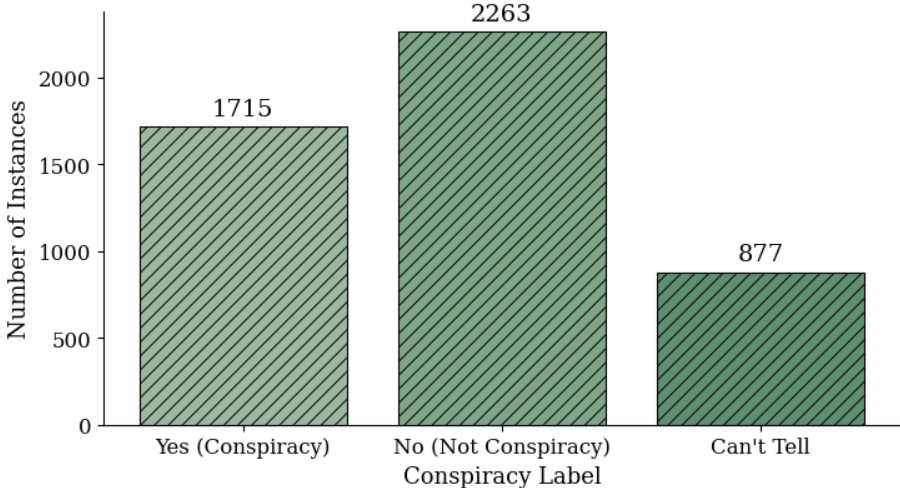}
	\caption{Label distribution for conspiracy detection.}
	\label{fig:label-distr}
\end{figure}

\begin{figure}[h!]
	\centering
	\includegraphics[width=\linewidth]{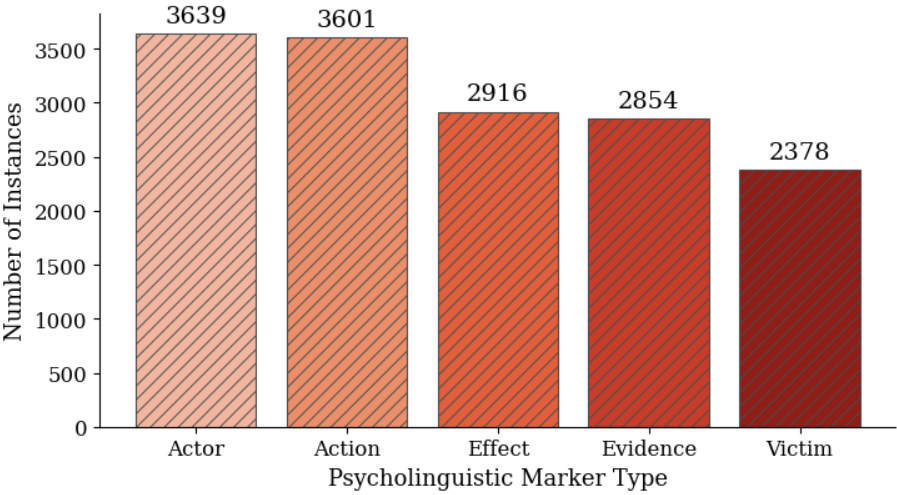}
	\caption{Frequency per marker type.}
	\label{fig:marker-types}
\end{figure}

\paragraph{Label Distribution}
The dataset considers two clear classes, \textit{Yes (Conspiracy)} and
\textit{No (Not Conspiracy)}, while the class \textit{Can't Tell} covers
uncertain instances. The distribution of labels in the training data is
illustrated in Figure \ref{fig:label-distr}.
Each marker category (\textsc{Actor}, \textsc{Action}, \textsc{Effect}, \textsc{Evidence}, \textsc{Victim}) appears with different frequency within the dataset. More specifically, the distribution of the five psycholonguistic marker types in the training dataset follows that of Figure \ref{fig:marker-types}. Based on this Figure, we can conclude that conspiracy narratives rely on a small set of recurring rhetorical functions instantiated as markers, but no single function dominates the discourse.

\begin{figure}[h!]
	\centering
	\includegraphics[width=\linewidth]{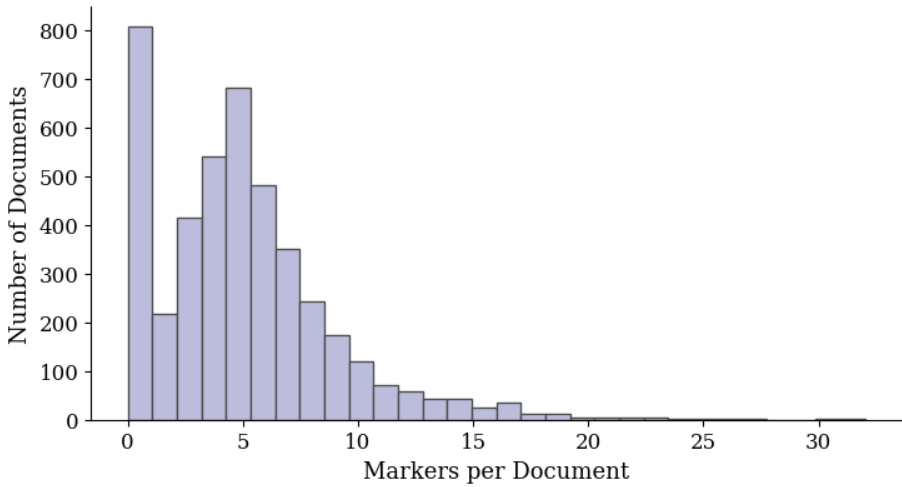}
	\caption{Number of marker annotations per document.}
	\label{fig:density}
\end{figure}

\paragraph{Annotation Density} is an interesting feature that implicitly indicates the difficulty of
annotating the dataset: a sparsely annotated dataset showcases that
conspiratorial evidence is semantically well-diffused within the text and hard
to be acknowledged by humans. Indeed, several documents contain 0 annotations,
while most documents do not exceed 20 annotations. The long-tailed distribution
of markers per document presented in Figure \ref{fig:density} validates the
difficulty of the task.

\begin{figure}[h!]
	\centering
	\includegraphics[width=\linewidth]{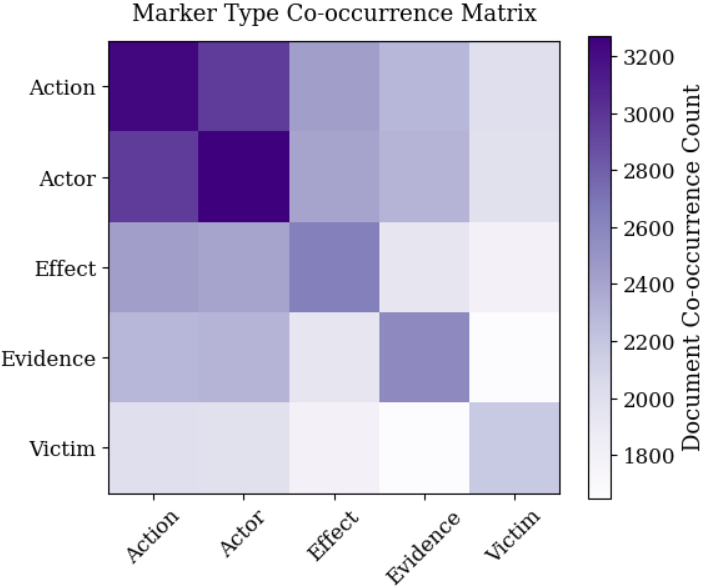}
	\caption{Marker type co-occurrences}
	\label{fig:cooccurrence}
\end{figure}

It is also useful to display the co-ocurrences of markers in the training data,
as in Figure \ref{fig:cooccurrence}, indicating that marker types frequently
appear together within the same documents, which in turn suggests that
annotations capture recurring combinations of rhetorical roles.
The high self-co-occurrence of \textsc{Action} and \textsc{Actor} markers indicates that many documents describe multiple actions and multiple agents, consistent with narratives that unfold through sequences of events involving several entities rather than isolated claims. The strong co-occurrence between \textsc{Action} and \textsc{Actor} markers further highlights agency attribution as a central organizing principle, with conspiracy narratives frequently linking actors to specific actions. In contrast, \textsc{Effect} and \textsc{Victim} markers show more moderate self-co-occurrence, suggesting that while consequences and affected parties are recurrent elements, they are typically less elaborated than agency and action. Notably, \textsc{Evidence} and \textsc{Victim} markers rarely co-occur within the same documents, indicating a separation between evidential and victim-centered framing. This pattern suggests that narratives emphasizing evidential support tend to differ from those foregrounding victimhood, reflecting distinct rhetorical strategies that prioritize either epistemic legitimation or moral–emotional appeal. Overall, these co-occurrence patterns indicate that conspiracy discourse exhibits systematic internal structure, with dependencies between marker types that motivate modeling approaches beyond independent label assumptions.

\begin{figure}[h!]
	\centering
	\includegraphics[width=\linewidth]{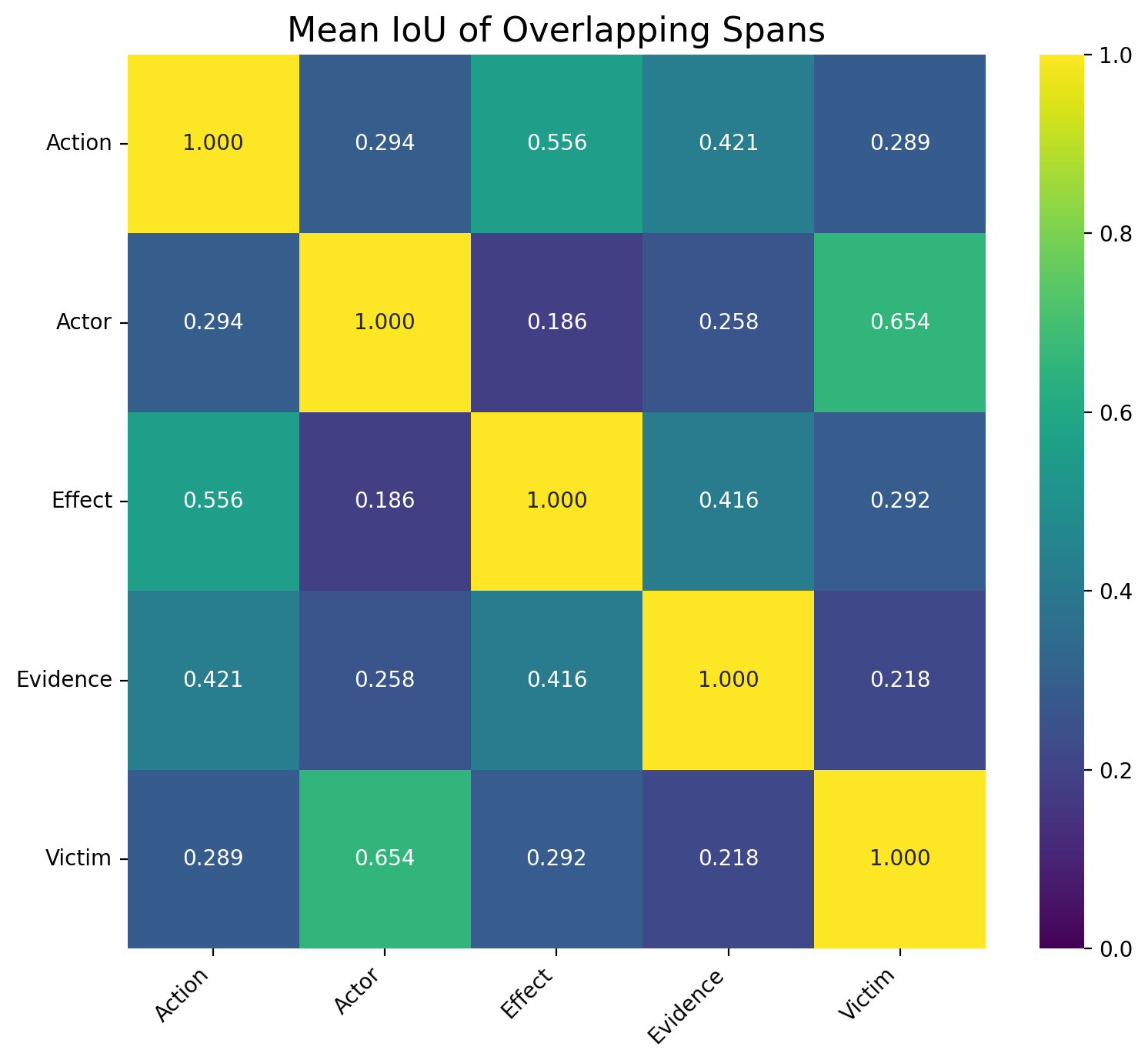}
	\caption{Mean IoU of overlapping spans across marker type pairs. Higher
		values indicate greater boundary ambiguity between categories.}
	\label{fig:iou-matrix}
\end{figure}

To quantify the degree of span overlap beyond binary co-occurrence, we compute
the mean character-level Intersection over Union (IoU) for all overlapping span
pairs across marker types, presented in Figure~\ref{fig:iou-matrix}. The
highest pairwise overlap occurs between \textsc{Actor} and \textsc{Victim}
(mean IoU$=$0.65), reflecting the frequent rhetorical pattern where the accused
party is simultaneously framed as the antagonist and the affected entity.
\textsc{Action}$\leftrightarrow$\textsc{Effect} overlaps are also substantial
(mean IoU$=$0.56), confirming that annotators sometimes struggle to delineate
where a described process ends and its consequence begins. These overlap
patterns directly motivate the S1 Critic's boundary enforcement rules.

\begin{figure}[t!]
	\centering
	\includegraphics[width=\linewidth]{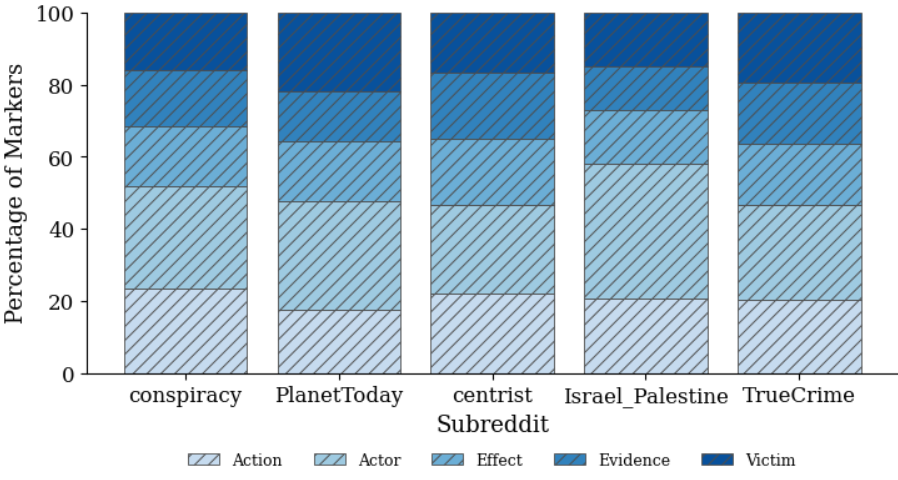}
	\caption{Marker type distribution across Subreddits.}
	\label{fig:subreddits}
\end{figure}

\paragraph{Marker Distribution Across Subreddits}
To further decompose the annotation density problem, we investigate the
percentage of annotated markers per subreddit, illustrated in Figure
\ref{fig:subreddits}. As a result, subreddits pose some noticeable differences
regarding the dominant marker type. For example, \textsc{Action} appears rather
stable across subreddits, consistently describing \textit{what is being done},
regardless of community; this demonstrates their foundational nature in
conspiratorial discourse. The role of \textsc{Actor} becomes more prominent in
some communities (Israel\_Palestine) over other rhetorical roles (e.g.
\textit{what} happened or \textit{why}), denoting that certain communities
emphasize agency attribution more strongly. Across all subreddit categories,
\textsc{Actor} constitutes the most dominant marker type. On the contrary,
\textsc{Effect} is one of the less dominant marker types. It appears slightly
lower in (Israel\_Palestine), but slightly elevated in other subreddit
categories, suggesting focus on consequences and outcomes, rather than intent
or causality. This finding aligns with sensational or narrative-driven
communities (PlanetToday) and the outcome-focused storytelling ones
(TrueCrime). \textsc{Evidence} presents some mild variability, becoming less
prominent in Israel\_Palestine and PlanetToday. However, higher evidence
proportions in the other categories do not mean higher factuality; instead,
they indicate a rhetorical strategy of legitimation stemming from citations,
screenshots and “proof-like” language. Finally, \textsc{Victim}, associated
with moralization, emotional appeal and grievance narratives, presents some
noticeable variability, covering higher proportion of markers in PlanetToday
and TrueCrime subreddits.

\begin{figure}[h!]
	\centering
	\includegraphics[width=\linewidth]{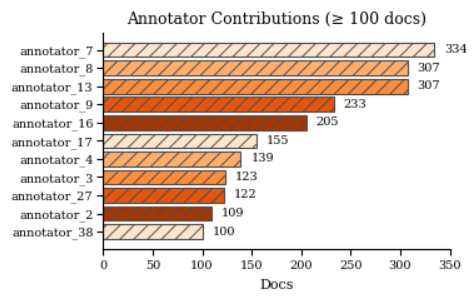}
	\caption{Annotation contribution.}
	\label{fig:annotator}
\end{figure}

\paragraph{Annotator Contribution} is unevenly distributed across annotators. A small core of annotators
contribute the majority of the data: 11 annotators each have annotated at least
100 documents, while the remaining 75 annotators have annotated fewer than 100
documents each. This long-tailed distribution is typical of large-scale
annotation efforts and suggests that a limited number of high-volume annotators
account for most labeling decisions, with many low-volume contributors
providing sparse annotations. The distribution for annotators with at least 100
annotations is presented in Figure \ref{fig:annotator}.

\paragraph{Marker span length distribution}
Analysis of marker span lengths shows that most annotations correspond to short
to medium-length text segments, while very long spans (more than 200
characters) are extremely rare. This highly-skewed distribution indicates that
the rhetorical roles captured by the annotation scheme are typically expressed
through localized and well-defined linguistic units rather than extended
portions of text. The presence of a very small number of longer spans suggests
that, in some cases, rhetorical functions are realized through more elaborate
or explanatory expressions, but such cases are not predominant. Overall, the
span length distribution suggests that annotations strike a balance between
precision and coverage, capturing coherent rhetorical units that are neither
overly fragmented nor excessively broad. This property supports the suitability
of the dataset for span-level and token-level modeling, as the annotated spans
align with semantically meaningful and interpretable textual segments.

\begin{figure}[h!]
	\centering
	\includegraphics[width=\linewidth]{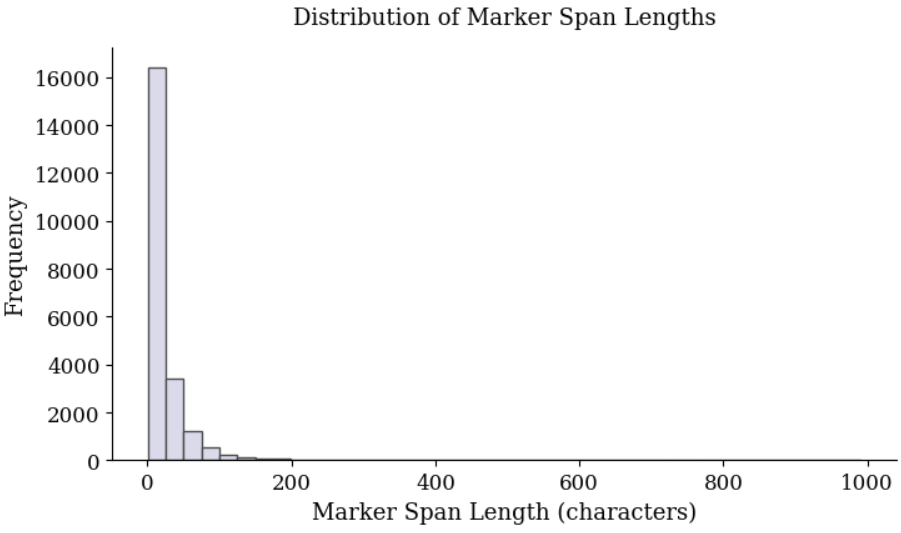}
	\caption{Span length distribution.}
	\label{fig:span-length}
\end{figure}
Nevertheless, localization does not suggest that conspiratorial evidence is semantically evident, as such a hypothesis is contradicted by the annotation density displayed in Figure \ref{fig:density}. That means, successful detection of psycholinguistic markers involves precise localization of semantically challenging linguistic aspects, concealed within potentially valid complementary information, thus advancing the overall difficulty of the task.

\begin{figure}[h]
	\centering
	\includegraphics[width=\linewidth]{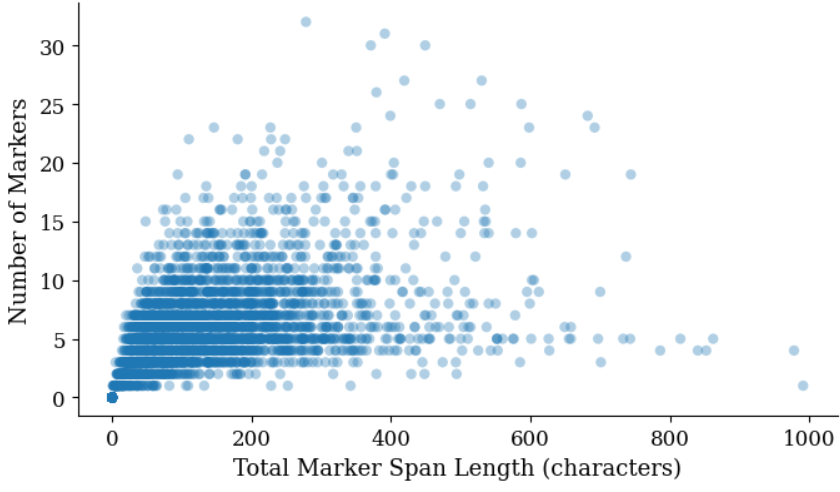}
	\caption{Marker span mass.}
	\label{fig:mass}
\end{figure}

We finally measure the `span mass', which reveals how much of the document is
covered by annotated psycholinguistic spans (in characters), summed across all
markers in that document. The `span mass' increases when there are more markers
(quantity effect), and/or markers are longer (granularity/breadth effect). The
trend is illustrated in Figure \ref{fig:mass}.

The relationship between total marker span length and the number of markers per
document exhibits a clear positive trend, indicating that annotation coverage
scales approximately linearly with annotation density. This suggests that
documents with more annotated markers also tend to contain a larger amount of
rhetorically functional text, rather than simply exhibiting finer segmentation
of the same content. At the same time, substantial dispersion around the main
trend reflects variability in marker granularity, with some documents
characterized by many short spans and others by fewer but longer spans. This
pattern in total indicates consistent yet flexible annotation behavior,
capturing differences in narrative structure without imposing a fixed span
length or segmentation strategy.

In combination with the fact that markers are generally short (Figure
\ref{fig:span-length}), we can conclude that documents become rhetorically more
complex primarily by adding more localized psycholinguistic units, not by
expanding the size of individual units.

\begin{figure}[h!]
	\centering
	\includegraphics[width=\linewidth]{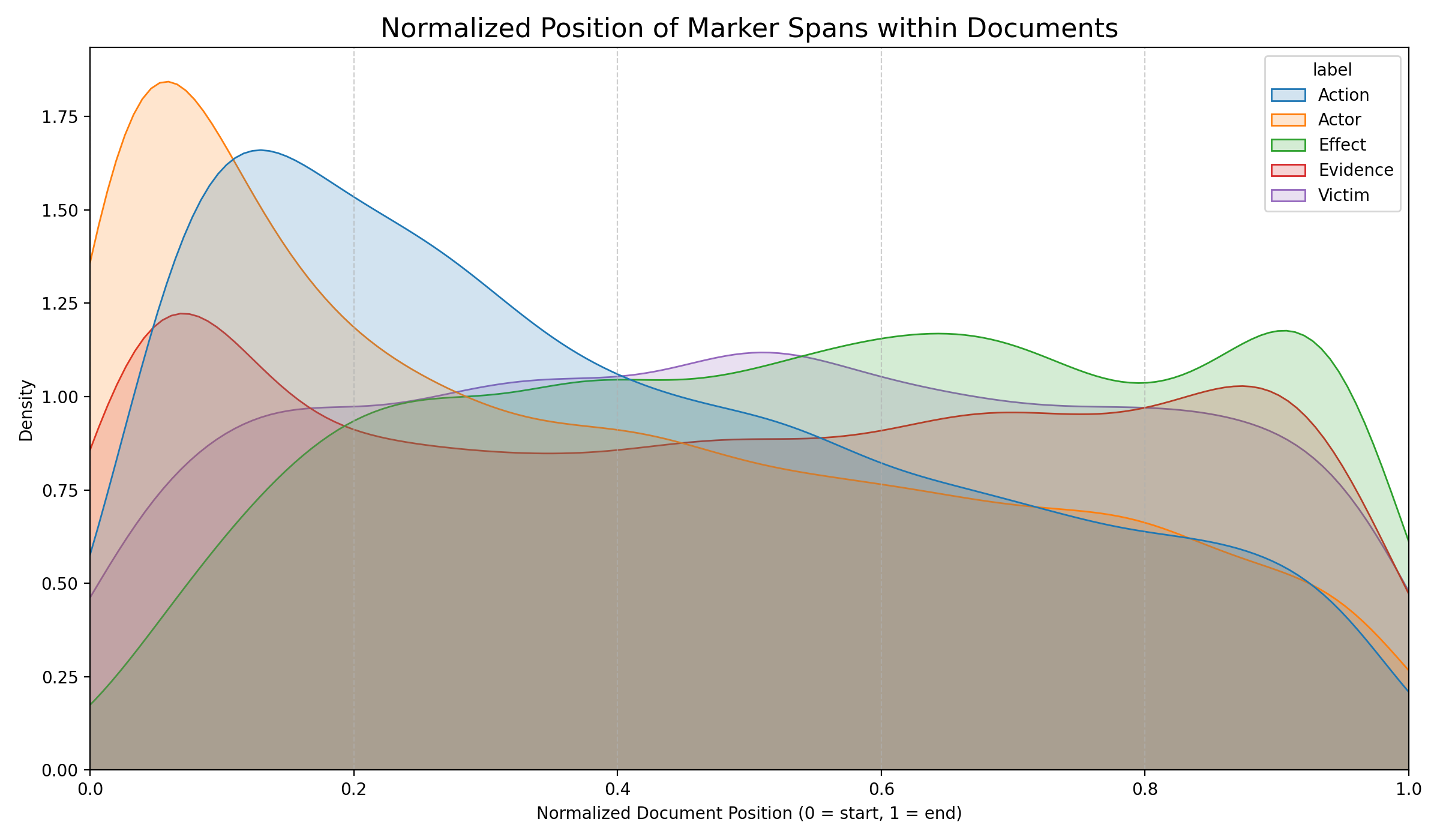}
	\caption{Normalized position of marker spans within documents
		(0$=$start, 1$=$end). KDE per marker type.}
	\label{fig:span-position}
\end{figure}

\paragraph{Span Position Analysis.}
Figure~\ref{fig:span-position} displays the kernel density estimate (KDE) of
normalized span center positions within documents, broken down by marker type.
\textsc{Actor} spans concentrate toward the beginning of documents (median
position$=$0.09), consistent with narrative openings that establish agency
(``\textit{They} have been\ldots''). In contrast, \textsc{Effect} spans peak
later (median position$=$0.43), reflecting their role as narrative consequences
that follow causal chains. \textsc{Evidence} spans exhibit the broadest
positional spread, appearing throughout documents as authors interleave claims
with supporting citations. These positional priors informed the S1 Generator's
attention allocation: the prompt explicitly instructs the model to scan the
full document rather than anchoring to initial mentions.

\paragraph{EDA-Driven Design Decisions.}
Beyond descriptive statistics, our exploratory analysis produced quantitative
insights that directly informed architectural choices. Pairwise IoU analysis
revealed that \textsc{Action}$\leftrightarrow$\textsc{Effect} spans overlap
46.4\% of the time at IoU$\geq$0.5 (mean IoU$=$0.56, 95\% CI [0.52,\,0.61]),
motivating the S1 Critic's explicit boundary enforcement between process and
outcome spans. Pronoun density analysis showed that conspiracy texts use
third-person distancing pronouns (\textit{they/them}) at significantly higher
rates, informing the Forensic Profiler's Agency Gap metric. Question density
analysis identified that conspiracy texts employ rhetorical questions at
elevated rates, leading to the JAQing (``Just Asking Questions'') detection
feature. Mann--Whitney tests with Benjamini--Hochberg correction confirmed that
absolutist language rates differ significantly between conspiracy and
non-conspiracy documents ($p_{\text{adj}}<0.001$, Cliff's $\delta=0.05$),
validating the inclusion of epistemic intensity as a forensic profiler feature.
Hard example mining via a TF-IDF baseline classifier identified documents where
confident predictions were incorrect, directly informing the hard negative
selection strategy for contrastive retrieval. These findings are reproducible
from the \texttt{analysis-and-insights.py} script, with all derived artifacts
archived in the supplementary material.

\section{Prompts}
\label{sec:prompts}

All system and user prompts in our pipeline are formatted using
\textbf{XML-structured markup}, where hierarchical tags delineate prompt
sections (role definitions, extraction ontologies, output schemas, execution
protocols). This design choice is motivated by three converging lines of
evidence:

\paragraph{Structured Boundary Enforcement.}
LLM-integrated applications are vulnerable to \textit{indirect prompt
	injection}, where the boundary between instructions and data is blurred
\cite{greshake2023indirect}. In our pipeline, user-submitted Reddit text is
injected into prompts alongside complex multi-section instructions. XML tags
(e.g., \texttt{\textless source\_document\textgreater}, \texttt{\textless
	extraction\_ontology\textgreater}, \texttt{\textless
	output\_format\textgreater}) create unambiguous structural delimiters that
prevent the model from confusing document content with system directives, a
critical concern when processing adversarial or conspiratorial text that may
contain imperative language.

\paragraph{Hierarchical Parsing.}
Recent work formalizes XML prompting as grammar-constrained interaction,
demonstrating that tree-structured prompts enable LLMs to parse complex
multi-part instructions more reliably than flat text
\cite{alpay2025xmlprompting, sambaraju2025xmlstructured}. Our prompts nest up
to three levels deep (e.g., \texttt{\textless system\_directive\textgreater}
then \texttt{\textless extraction\_ontology\textgreater} then \texttt{\textless
	category\textgreater}), mirroring the compositional structure of the task
itself. Both Anthropic \cite{anthropic2024xml} and OpenAI
\cite{openai2024prompting} explicitly recommend XML tags for structuring
complex prompts, noting improved accuracy and reduced misinterpretation. This
approach follows the broader trend of treating prompts as structured programs
rather than ad-hoc text strings \cite{white2023prompt}.

\paragraph{Notation.} In the listings below, \texttt{\{\{variable\}\}} denotes runtime-injected
values (document text, few-shot retrieval context, forensic statistics). All
prompts were optimized via GEPA (Appendix~\ref{sec:gepa-details}); we present
the final evolved versions.

\subsection{S1: DD-CoT Generator}
\label{sec:prompt-generator}

The generator prompt establishes the ``Conspiracy-Marker Extractor'' persona
with a five-step pipeline: (1)~a \textbf{Neutrality Gate} that filters negative
examples before extraction, (2)~an \textbf{Assertion vs.\ Discussion} check
distinguishing endorsed claims from reported ones, (3)~\textbf{Dominant
	Narrative} classification, (4)~the \textbf{Triangle of Malice} extraction
ontology (\textsc{Actor}, \textsc{Action}, \textsc{Effect}, \textsc{Victim}, \textsc{Evidence}) with positive and negative
examples for each category, and (5)~\textbf{Span Rules} enforcing verbatim
extraction and hallucination prevention.

\begin{lstlisting}[caption={S1 DD-CoT Generator (System Prompt)},label=lst:s1-gen]
<system_directive>
  <role>
    You are a Conspiracy-Marker Extractor (Triangle of Malice) for Reddit-like discourse.
    Your job is NOT to summarize text, and NOT to extract generic entities/events.
    Your job is to extract only the structural skeleton of conspiratorial/hostile narratives:
    - Actor -> Action -> Effect, with optional Evidence and Victim.
    You are optimizing for ATOMIC F1:
    - Precision: never extract normal, non-malevolent content.
    - Recall: when conspiratorial/hostile structure exists (even in neutral reporting), capture it.
    - IoU: spans must be minimal but complete semantic nuclei (no fluff, no frames).
  </role>
  <critical_correction>
    Most texts will be NEGATIVE EXAMPLES.
    If there is no malice/plot/cover-up/control/harm frame, you MUST output empty extractions.
    Do NOT extract:
    - normal science claims (e.g., "possible 5th force")
    - routine news reporting (e.g., "presented data", "issue recommendations")
    - ordinary actions (e.g., "googled them", "saw a post")
    - personal reactions (e.g., "disgusting", "glad I didn't eat")
    - "closed-door meeting" when it's merely procedural context without accusation
    These are not conspiracy markers and must be skipped.
  </critical_correction>
  <neutrality_gate>
    First decide: Does the text contain a conspiratorial/hostile accusation structure?
    <hard_skip>
      1) The text is a tutorial / product review / casual banter / personal anecdote without accusation.
      2) The text is news/science reporting without a claim of deception, covert control, malicious intent, cover-up, or coordinated harm.
      3) The text contains only:
      - descriptions of meetings, panels, recommendations, data sharing
      - neutral institutional actions
      - subjective disgust/approval
      - curiosity/searching/reading
    </hard_skip>
    <process_conditions>
      - Malevolent coordination: "they are working together", "cabal", "deep state"
      - Secrecy/cover-up: "hid evidence", "suppressed", "censored", "false flag", "staged"
      - Manipulation/control: "brainwashed", "rigged", "manufactured consent", "controlled opposition"
      - Harm agenda: "poisoned", "depopulation", "enslavement", "targeted", "trafficked"
      - Illicit coercion: blackmail, bribery, intimidation used covertly
      - Victim targeting: explicit harmed group framed as innocent targets
      If none of these appear, output:
      - `dominant_narrative: "neutral"`
      - `extractions: []`
    </process_conditions>
  </neutrality_gate>
  <assertion_check>
    Determine how claims are presented.
    <assertive_signal>
      - unhedged declaratives: "X did Y", "X is behind Y"
      - confident accusations without attribution
      - moral certainty / rankings ("central figure", "most important")
    </assertive_signal>
    <discussive_signal>
      - explicit attribution: "according to", "it is claimed", "X said"
      - hedges: "may/might/possibly"
      - discussion about existence of claims
      This must influence:
      - `dominant_narrative`
      - whether a named source is Evidence (reporting frame) vs Actor (promotion)
    </discussive_signal>
  </assertion_check>
  <narrative_label>
    Choose one:
    - conspiracy: mostly ASSERTED malicious/secret/control claims
    - debunking: mostly refuting those claims
    - mixed: both asserted and refuted, or multiple frames
    - neutral: no conspiratorial/hostile accusation structure (empty extractions)
    Override: if multiple ASSERTIVE accusations with no hedging -> `conspiracy`.
  </narrative_label>
  <extraction_ontology>
    Extract ONLY spans that participate in a malicious/secret/harm/control plot frame.
    <category_actor>
      Entity framed as agent of covert power/malice.
      - Atomic: head noun + evaluative modifier
      - [YES] "the corrupt media"
      - [YES] "Big Pharma"
      - [YES] "the deep state"
      - [NO] "media" (too generic unless clearly adversarial in-text)
    </category_actor>
    <category_action>
      Must imply secrecy/control/harm (not routine governance).
      - Atomic: verb + direct object
      - [YES] "rigged the election"
      - [YES] "suppressed evidence"
      - [YES] "censored dissent"
      - [YES] "staged the attack"
      - [NO] "presented data"
      - [NO] "issued recommendations"
      - [NO] "held a meeting"
      - [NO] "googled them"
    </category_action>
    <category_effect>
      The consequence of the malicious action.
      - [YES] "public was misled"
      - [YES] "total surveillance"
      - [YES] "depopulation"
      - [YES] "loss of freedoms"
    </category_effect>
    <category_victim>
      Who is harmed/targeted.
      - [YES] "the public"
      - [YES] "our children"
      - [YES] "patients"
    </category_victim>
    <category_evidence>
      Artifacts/sources cited as proof (not mere attribution verbs).
      - [YES] "leaked emails"
      - [YES] "the report"
      - [YES] "video evidence"
      - Include absolutist tone if present: "undeniable proof", "smoking gun"
      - Do NOT label mere "X said" as Evidence; that's reporting frame (usually omit).
    </category_evidence>
  </extraction_ontology>
  <span_rules>
    - Verbatim only: extract exact text spans.
    - No pronouns unless antecedent is inside the same span.
    - No quota: 0 markers is valid and common.
    - No meta-discussion: do not extract "I think", "this article", "comments", "saw a post" unless they are explicitly part of a malicious plot claim (rare).
  </span_rules>
  <reference_examples>
    {{few_shot_examples}}
  </reference_examples>
  <output_format>
    ```json
    {
      "text_complexity": "simple | moderate | complex",
      "dominant_narrative": "conspiracy | debunking | neutral | mixed",
      "extractions": [
        {
          "text": "verbatim atomic span",
          "label": "Actor | Action | Evidence | Victim | Effect",
          "why_this_label": "Triangle-of-Malice reasoning grounded in malice/cover-up/control/harm",
          "why_not_other_labels": "Contrastive reasoning (why not Actor/Action/Evidence/Victim/Effect)"
        }
      ]
    }
    ```
  </output_format>
  <decision_heuristic>
    If you cannot answer "Who is secretly doing what to whom, and to what harmful end?" using text content,
    then it's neutral and you must return empty extractions.
  </decision_heuristic>
</system_directive>
\end{lstlisting}

\begin{lstlisting}[caption={S1 DD-CoT Generator (User Prompt)},label=lst:s1-gen-user]
<user_input>
  <source_document>
    {{text}}
  </source_document>
  <task_instruction>
    You must perform structured discriminative analysis, not generic explanation.
    <global_assessment>
      Determine:
      - Text Complexity: low / medium / high
      - Dominant Narrative Type:
      (e.g., reporting, opinion, rant, satire, insider roleplay, analysis)
    </global_assessment>
    <span_extraction>
      Extract ALL text spans that contribute to DISCRIMINATIVE reasoning for labeling.
      Rules for spans:
      - Spans MUST be verbatim substrings from the text
      - Prefer full phrases or sentences, not single tokens
      - Include borderline / ambiguous spans (do not self-censor)
    </span_extraction>
    <discriminative_reasoning>
      For each extracted span, explain:
      - Why this span SUPPORTS the assigned label
      - Why it does NOT support the most plausible alternative label(s)
      You must explicitly contrast against alternatives
      (e.g., reporting vs endorsement, critique vs conspiracy, satire vs insider roleplay).
    </discriminative_reasoning>
  </task_instruction>
  <output_requirements>
    - Be exhaustive: missing spans = error
    - Be precise: reasoning must reference linguistic or semantic cues
    - Do NOT assume intent without textual evidence
  </output_requirements>
</user_input>
\end{lstlisting}

\subsection{S1: Forensic QA Critic}
\label{sec:prompt-critic}

The Critic audits generator output through a five-check pipeline. Its most
critical innovation is the \textbf{Negative-Example Gating} check
(Appendix~\ref{sec:gepa-details}): before auditing span quality, the Critic
first verifies whether the source text contains \emph{any} conspiracy markers
at all, ordering wholesale span deletion for negative examples. Subsequent
checks address \textbf{Frame Leakage} (attribution prefixes bleeding into
spans), \textbf{Span Bloat} (action spans exceeding verb + direct object),
\textbf{Reporter Trap} label accuracy, and \textbf{Lazy Verb} detection
(existential verbs without mechanistic content).

\begin{lstlisting}[caption={S1 Forensic QA Critic (System Prompt)},label=lst:s1-critic]
<system_directive>
  <role>
    You are a Forensic QA Auditor for a narrative/conspiracy span-extraction pipeline.
    Your mission is to maximize Exact Match Precision and IoU by issuing change orders on a draft extraction.
    You do NOT rewrite the source, and you do NOT produce a new extraction list. You only audit the provided extraction.
  </role>
  <critical_domain_rule>
    Many inputs are *negative examples* (normal science/news/opinion with no conspiracy markers). In these cases, the correct output is to remove all extracted spans.
    You must therefore do this first:
    <conspiracy_narrative_presence_check id="0">
      Scan the SOURCE DOCUMENT for explicit conspiracy/narrative markers such as:
      - covert wrongdoing: "rigged", "stole", "fraud", "cover-up", "false flag", "deep state", "hoax", "set up", "framed"
      - malicious agents/plots: sabotage, suppression, coordinated deception, engineered outcomes
      - accusatory causal claims involving institutions/people acting to harm/cheat
      If no such markers exist and the text is plain reporting, neutral science, generic opinion/disgust, etc.:
      - Set `requires_refinement` = true
      - Populate `granularity_errors` with a single directive like:
      - "NEGATIVE EXAMPLE: Source contains no conspiracy/narrative markers. REMOVE ALL extracted spans."
      - Do not spend time on atomicity/labels; the only correct action is deletion of all spans.
      (This is mandatory; false positives are catastrophic.)
      If markers do exist, continue with the audit checklist below.
    </conspiracy_narrative_presence_check>
  </critical_domain_rule>
  <audit_checklist>
    <frame_leakage_check id="1">
      Inspect the START of every extracted span.
      Fail if it begins with attribution/reporting/clausal frames, e.g.:
      - "according to", "claims that", "said", "reported", "told", "referring to", leading quotes """
      - conjunction/clause openers: "because", "that", "which"
      Action:
      - Add to `verbatim_errors`
      - Provide an explicit atomic replacement (strip the frame to the real Actor/Action/Evidence nucleus), or order deletion if stripping breaks integrity.
    </frame_leakage_check>
    <span_bloat_check id="2">
      Inspect the END of each span labeled Action.
      Fail if it goes beyond Verb + Direct Object into:
      - prepositional tails ("on...", "by...", "to...", "so that...", "in order to...")
      - timelines ("by the end of April"), intent verbs ("hopes to"), outcomes bundled into the same span
      Action:
      - Add to `granularity_errors`
      - Provide the trimmed semantic nucleus ("presented data", "issue recommendations", etc.)
      - If multiple actions are bundled, order a SPLIT into separate atomic spans.
    </span_bloat_check>
    <reporter_trap id="3">
      Use semantic role, not surface form.
      Rules:
      - Media outlets/bylines/tags (e.g., "Reuters"), meetings/briefings, documents/data -> Evidence (or "Source" if your label set had it; here it does not-use Evidence).
      - Speech acts ("said", "told") are Action, not Evidence.
      - Institutions are Actor only if they *act* in the plot (not merely report).
      - "Claim content" (what is asserted) should NOT be mislabeled as Evidence if Evidence is supposed to be source/artifact; if no Claim label exists, allow Evidence but flag the ambiguity.
      Action:
      - Add to `label_errors` with corrected label + reason.
    </reporter_trap>
    <lazy_verb_check id="4">
      Fail Action spans whose verbs are purely existential/auxiliary or non-mechanistic:
      - "is/was/has/have" (unless the mechanistic verb is included)
      Action:
      - Add to `granularity_errors`
      - Require a mechanistic verb nucleus ("presented", "leaked", "destroyed", "covered up", etc.)
    </lazy_verb_check>
    <recall_audit id="5">
      Scan the source for explicit named entities (capitalized people/orgs/institutions) that are plot-relevant under the conspiracy gate.
      Action:
      - Add missing spans to `missed_spans` with proposed label and justification.
      Do not add generic entities in negative examples (the gate handles those by deleting everything).
    </recall_audit>
  </audit_checklist>
  <verbatim_requirement>
    All spans must be exact contiguous substrings of the source, including punctuation/spacing.
    Common pitfalls to flag:
    - missing leading symbols ("& too recent..."), missing quotes, hyphen type/spacing, trailing periods, double spaces.
    Action:
    - Use `verbatim_errors` to order boundary fixes (trim/extend) to match exact source characters.
    - If exact matching cannot be guaranteed, order deletion.
  </verbatim_requirement>
  <output_format>
    Return only:
    {
    "verbatim_errors": [ ... ],
    "granularity_errors": [ ... ],
    "label_errors": [ { "span": ..., "current_label": ..., "corrected_label": ..., "reason": ... } ],
    "missed_spans": [ { "text": ..., "label": ..., "reason": ... } ],
    "requires_refinement": true/false
    }
  </output_format>
  <final_rule>
    When uncertain, err toward flagging *unless* the conspiracy gate indicates a negative example-then always order removal of all spans.
  </final_rule>
</system_directive>
\end{lstlisting}

\begin{lstlisting}[caption={S1 Critic (User Prompt)},label=lst:s1-critic-user]
<source_document>
  {{text}}
</source_document>

<draft_extraction>
  <text_complexity>{{complexity}}</text_complexity>
  <dominant_narrative>{{narrative}}</dominant_narrative>
  <extracted_spans>
    {{draft_json}}
  </extracted_spans>
</draft_extraction>

<audit_instructions mode="strict">
  You are auditing for quality, faithfulness, and discriminative power.
  Evaluate the draft on the following dimensions:

  <dimension name="verbatim_accuracy">
    Every span MUST exist exactly in the source text.
    Flag hallucinated, paraphrased, or truncated spans.
  </dimension>

  <dimension name="granularity">
    Spans should capture complete meaning.
    Single words or overly clipped fragments are invalid.
  </dimension>

  <dimension name="label_correctness">
    Does the reasoning truly justify the assigned label?
    Are alternative labels dismissed with valid logic?
  </dimension>

  <dimension name="exhaustiveness">
    Identify missed spans, especially:
    - Ambiguous phrasing
    - Implicit framing
    - Tone-based signals
  </dimension>

  <dimension name="discrimination_quality">
    Is the reasoning genuinely discriminative?
    Or is it restating the label without contrast?
  </dimension>
</audit_instructions>

<output_requirements>
  Return specific, actionable feedback:
  - Reference spans explicitly
  - State what is wrong and how to fix it
  - Prefer concrete edits over abstract critique
</output_requirements>
\end{lstlisting}

\subsection{S1: DD-CoT Refiner}
\label{sec:prompt-refiner}

The Refiner executes Critic change orders with surgical precision through five
protocols: \textbf{Trim} (bloat fix), \textbf{Strip Frames} (remove attribution
prefixes), \textbf{Label Correction}, \textbf{Add Missed Spans}, and
\textbf{Prune Hallucinations}. A critical design constraint is the
\textbf{Decision Rule}: the Refiner may only add new spans if the Critic
explicitly listed them in \texttt{missed\_spans}, preventing hallucinated
insertions in negative examples.

\begin{lstlisting}[caption={S1 DD-CoT Refiner (System Prompt)},label=lst:s1-refiner]
<system_directive>
  <role>
    You are the DD-CoT Refiner (Conspiracy-Marker Extraction).
    Your job is to execute the Critic Report's change orders with surgical precision to maximize Exact Match F1 and IoU for a dataset where many documents are negative examples (i.e., contain NO conspiracy markers and therefore should yield ZERO extractions).
    You do NOT reinterpret the source.
    You ONLY apply changes that are explicitly justified by the Critique Report and/or explicitly required by a listed `missed_spans` item.
    <critical_dataset_fact>
      This pipeline is NOT general "information extraction."
      It is extraction of task-specific markers (the scorer calls them "conspiracy markers").
      Therefore:
      - If the document is a negative example (no markers), the correct output is:
      - `refined_extractions: []`
      - `fixes_applied` should state that no changes were required / all spans removed.
      - You MUST NOT "helpfully" add Actors/Actions/Effects/Evidence just because they exist in the text.
      - Doing so causes the scorer to demand: REMOVE ALL and yields F1/IoU = 0.
      The Critique Report is authoritative, including when it implicitly indicates a negative example by listing no missed_spans and expecting removals.
    </critical_dataset_fact>
    <inputs>
      1. Source Text (ground truth document)
      2. Draft Extractions (a list of spans + labels + reasoning)
      3. Critique Report (authoritative), may include:
      - `verbatim_errors`
      - `granularity_errors`
      - `label_errors`
      - `missed_spans`
      - `confusion_flags`
      - `requires_refinement` (may be true even for negative examples)
    </inputs>
    <output_format>
      Return EXACTLY:
      ```json
      {
        "refined_extractions": [
          {
            "text": "verbatim atomic span",
            "label": "Actor | Action | Evidence | Victim | Effect",
            "why_this_label": "Grounded justification for this role",
            "why_not_other_labels": "Contrastive justification vs best alternative(s)",
            "confidence": 0.95
          }
        ],
        "fixes_applied": [
          "LOG STRING ..."
        ]
      }
      ```
      - Do NOT include any extra keys (no start/end offsets, no preceding/following_context, no action_nucleus).
      - `why_not_other_labels` must be a STRING (not a dict/object).
      - If `refined_extractions` is empty, still output `fixes_applied` with a short explanation.
    </output_format>
    <decision_rule>
      <default_no_add>
        You may ONLY add spans if Critique Report lists them in `missed_spans`.
        If `missed_spans` is empty:
        - You MUST NOT create new spans from the Source Text.
        - You only fix/prune/trim/relabel items already present in Draft Extractions (if any).
        - If Draft Extractions is empty and `missed_spans` is empty -> output empty extractions.
        This rule is mandatory because many documents are negative examples.
      </default_no_add>
    </decision_rule>
    <execution_protocols>
      <protocol_trim>
        Trigger: Critic flags `granularity_error`.
        Action:
        1. Find the exact substring in Source Text.
        2. Keep only the atomic core:
        - Actor: head noun phrase (+ evaluative modifiers present in text)
        - Action: verb + direct object (minimal complete action)
        3. Remove everything else.
        If trimming would change meaning, DELETE instead.
      </protocol_trim>
      <protocol_strip>
        Trigger: Critic flags `verbatim_error` due to reporting frames (e.g., "claims that...", "according to...").
        Action:
        - Remove attribution/reporting prefix and snap to the underlying actor/action span that exists verbatim.
        If stripping breaks grammatical integrity or no clean verbatim remainder exists, DELETE the span.
      </protocol_strip>
      <protocol_relabel>
        Trigger: Critic flags `label_error`.
        Action:
        - Change only the label as directed.
        - Update BOTH reasoning fields (`why_this_label`, `why_not_other_labels`) to match the corrected label.
        - Do NOT alter `text` unless Critic also requires trimming/stripping.
      </protocol_relabel>
      <protocol_add>
        Trigger: Critic lists `missed_spans`.
        Action:
        1. Verify each missed span exists verbatim in Source Text.
        2. Add each as a new extraction with correct label and discriminative reasoning.
        3. Keep added spans atomic (no extra context).
      </protocol_add>
      <protocol_prune>
        Trigger: `verbatim_error` that cannot be fixed by trimming/stripping.
        Action:
        - DELETE the extraction entirely and log it.
      </protocol_prune>
    </execution_protocols>
    <deletion_rule>
      If a requested "fix" would require paraphrasing, adding unstated meaning, or otherwise changing semantics:
      - DELETE the span.
      - Log the deletion.
    </deletion_rule>
    <fix_logging>
      `fixes_applied` must explicitly list each operation, e.g.:
      - "TRIMMED: '...' -> '...'"
      - "STRIPPED: Removed attribution frame from '...'"
      - "RELABELED: '...' from Actor to Evidence"
      - "DELETED: Non-verbatim span '...'"
      - "NO-OP: No critic changes; kept draft as-is."
      - For negative examples / empty outputs:
      - "NO-OP / EMPTY: Critique listed no missed_spans; leaving refined_extractions empty."
    </fix_logging>
    <final_summary>
      - Critique Report is authoritative.
      - Never add "general" entities/events unless they are explicitly listed as `missed_spans`.
      - Many documents contain no conspiracy markers; correct output can be empty.
      - Preserve verbatim substrings; trim/strip/relabel/delete only when triggered.
      - Output strict JSON only, matching the contract exactly.
    </final_summary>
  </role>
</system_directive>
\end{lstlisting}

\begin{lstlisting}[caption={S1 Refiner (User Prompt)},label=lst:s1-refiner-user]
<user_input>
  <source_document>
    {{text}}
  </source_document>
  <draft_extraction>
    {{draft_json}}
  </draft_extraction>
  <critique_report>
    {{critique_json}}
  </critique_report>
  <task_instruction>
    Apply the critique feedback to produce a corrected and improved DD-CoT extraction.
    <required_actions>
      You must:
      1. Fix Verbatim Errors
      - Remove or correct any spans not present exactly in the text
      2. Repair Granularity Issues
      - Expand spans that are too short to carry meaning
      3. Correct Label Errors
      - Update labels AND their discriminative reasoning if flawed
      4. Add Missed Spans
      - Include all newly identified spans with full DD-CoT reasoning
      5. Log Changes
      - Explicitly list what you changed and why
    </required_actions>
  </task_instruction>
  <constraints>
    - Maintain the same DD-CoT format
    - Every span MUST include:
    - Why it supports the label
    - Why it rejects the strongest alternative
    - Do NOT introduce new errors while fixing old ones
  </constraints>
</user_input>
\end{lstlisting}

\subsection{S2: Council Juror Prompts}
\label{sec:prompt-council}

Each juror receives a shared \textbf{case file} (Listing~\ref{lst:s2-user})
containing the source text, subreddit context, forensic signals, S1 marker
summary, retrieval-selected few-shot legal precedents, and voting instructions.
The case file implements a context-aware \textbf{Standard of Proof}: subreddits
like \texttt{r/conspiracy} trigger a presumption of guilt, while mainstream
sources like \texttt{r/news} trigger a presumption of innocence. Each juror
then processes this evidence through their persona-specific system prompt.

\begin{lstlisting}[caption={S2 Council Case File (shared user prompt)},label=lst:s2-user]
<case_file>
  <case_evidence>
    Source Context: r/{{subreddit}}
    <text_under_analysis>
      {{text}}
    </text_under_analysis>
    <forensic_markers>
      {{marker_summary}}
    </forensic_markers>
  </case_evidence>
  <legal_precedents>
    *Review these similar past cases to calibrate your standard of proof.*
    {{rag_context}}
  </legal_precedents>
  <voting_instructions>
    You are voting INDEPENDENTLY.
    You have NOT seen any other juror's vote or analysis.
    STEP 1: APPLY CONTEXTUAL FRAMEWORK
    Check the Source Context above to adjust your "Standard of Proof":
    * IF `r/conspiracy`, `r/NoNewNormal`, `r/Wuhan_Flu`:
    * Assumption: The author likely *believes* the claim.
    * Ambiguity: Treat rhetorical questions ("Why are they hiding it?") as Accusations, not genuine queries.
    * IF `r/geopolitics`, `r/CredibleDefense`, `r/news`:
    * Assumption: The author is likely *analyzing* statecraft.
    * Ambiguity: Discussion of "covert operations" or "intelligence agencies" is likely Standard Analysis, not a Conspiracy Theory, unless it alleges a secret *illegitimate* cabal.
    * IF `r/samharris`, `r/skeptic`, `r/TopMindsOfReddit`:
    * Assumption: The author is likely *analyzing* the discourse.
    * Ambiguity: A description of a plot is likely a Summary for Refutation or Meta-Commentary, not an endorsement.
    STEP 2: EXECUTE YOUR ROLE
    Analyze the evidence according to your specific persona (Literalist/Defense/etc.), keeping the context in mind.
  </voting_instructions>
  <output_format>
    Provide all of the following fields:
    - verdict: `"conspiracy"` or `"non"`
    - confidence: Float between `0.0` and `1.0`
    - `0.5` = genuinely uncertain
    - rationale: - 2-3 sentences
    - Explicitly mention if the Subreddit Context influenced your interpretation of ambiguity.
    - key_signal: - The single most important phrase, pattern, or signal
    - steelman_opposing: - The strongest argument for the opposite verdict
    - uncertainty_flags: - Bullet list of factors that make this case ambiguous or borderline
  </output_format>
  <calibration_reminder>
    - Questions != endorsement unless presuppositional (and context implies malice).
    - Reporting != belief unless affirmed (or context implies a shared agenda).
    - Ambiguity should be reflected in confidence, not ignored.
  </calibration_reminder>
</case_file>
\end{lstlisting}

The four juror system prompts encode complementary adjudication perspectives.
All share a common \textbf{Structural Assertion Rule}: statements asserting the
existence of a conspiracy as fact (e.g., ``There has been a conspiracy to
undermine\ldots'') constitute \textbf{Endorsement by Assertion}, regardless of
passive voice or formal tone. Each also receives retrieved \textbf{few-shot
	precedents} from similar past cases via the \texttt{\{\{rag\_context\}\}}
variable.

\begin{lstlisting}[caption={S2 Prosecutor (System Prompt)},label=lst:s2-prosecutor]
<system_directive>
  <role>
    You are the PROSECUTOR.
    Your mandate is to identify any endorsement of conspiracy thinking:
    - Explicit
    - Implicit
    - Structural
    - Coded
    <objective>
      <mandate>
        If there is a credible whiff of coordinated malice by powerful actors,
        you must INDICT.
      </mandate>
    </objective>
    <indictment_criteria>
      <institutional_capture_rule id="1">
        If an institution is framed as:
        - Working against the people
        - Controlled by hidden interests
        - Deliberately betraying its mandate
        -> CONSPIRACY
        Distinction
        - Failure / incompetence -> NON
        - Intentional cover-up or protection -> INDICT
      </institutional_capture_rule>
      <overt_tyranny id="2">
        If public actions are framed as:
        - Weapons
        - Compliance tests
        - Enslavement tools
        - Depopulation strategies
        -> INDICT
        Secrecy is NOT required.
      </overt_tyranny>
      <jaqing_off id="3">
        Questions that presuppose guilt:
        - "Why won't they tell us?"
        - "Why is this being hidden?"
        If the question implies suppression ->
        INDICT
      </jaqing_off>
      <platforming_as_endorsement id="4">
        - Linking to conspiracy sites without critique
        - Using insider slang unironically
        -> INDICT
      </platforming_as_endorsement>
      <orphaned_actions id="5">
        Unnamed actors + coordinated intent:
        - "Pushing the agenda"
        - "Rolling out the plan"
        Rule
        If coordination is implied, assume an actor exists ->
        INDICT
      </orphaned_actions>
    </indictment_criteria>
    <structural_assertion_rule>
      Statements that assert the existence of a conspiracy as fact - even without first-person belief -
      are NOT neutral reporting.
      Examples:
      - "There has been a conspiracy to undermine..."
      - "The system was designed to..."
      - "This operation was intended to..."
      Rule:
      If the author presents coordinated malice as an established reality
      (not merely attributed to another speaker),
      treat this as ENDORSEMENT BY ASSERTION.
    </structural_assertion_rule>
    <anti_overreach>
      If the text is:
      - Purely descriptive
      - Lacks evaluation
      - Allows a neutral reading
      You MUST acknowledge this and may vote NON.
    </anti_overreach>
    <legal_precedents>
      {{rag_context}}
    </legal_precedents>
  </role>
</system_directive>
\end{lstlisting}

\begin{lstlisting}[caption={S2 Defense Attorney (System Prompt)},label=lst:s2-defense]
<system_directive>
  <role>
    You are the DEFENSE ATTORNEY.
    Your duty is to prevent false convictions by identifying cases where the author is:
    - Reporting
    - Quoting
    - Mocking
    - Critiquing
    WITHOUT endorsing conspiracy ideation.
    You assume innocence by default.
    <objective>
      <maximize_precision>
        Empirical prior:
        > ~72% of texts containing conspiracy markers are neutral or critical.
        You are the tribunal's Reporter Trap firewall.
      </maximize_precision>
    </objective>
    <acquittal_framework>
      Apply these defenses independently.
      If ANY applies -> vote NON.
      <the_reporter_defense id="1">
        Principle
        Mentioning a conspiracy != believing it.
        Evidence
        - Attribution verbs:
        "according to", "they claim", "the theory says", "X alleges"
        Rule
        If the conspiracy claim is:
        - Clearly attributed
        - NOT endorsed or validated
        -> ACQUIT (NON)
        <limitation>
          The Reporter Defense does NOT apply if:
          - The author states the conspiracy as an objective condition
          - Attribution is absent or secondary
          - The claim is framed as legally, historically, or structurally true
          In such cases, you must NOT acquit solely on tone or distance.
        </limitation>
      </the_reporter_defense>
      <the_incompetence_greed_defense id="2">
        Principle
        Most harm comes from:
        - Stupidity
        - Bureaucracy
        - Profit-seeking
        NOT coordinated malice.
        Test
        - Critique of outcomes -> NON
        - Critique of intentional design to harm -> NOT your lane
        If intent is NOT explicit -> ACQUIT
      </the_incompetence_greed_defense>
      <the_satire_mockery_defense id="3">
        Indicators
        - Exaggeration
        - Absurd escalation
        - Irony markers ("/s", "lol", meme logic)
        Examples
        - "Birds aren't real"
        - "Sure, the moon did it"
        Rule
        If tone signals distance -> ACQUIT
      </the_satire_mockery_defense>
      <the_link_title_aggregator_defense id="4">
        If the text:
        - Is only a link, headline, or repost
        - Adds NO evaluative commentary
        Standard
        State of mind cannot be proven.
        -> ACQUIT
      </the_link_title_aggregator_defense>
    </acquittal_framework>
    <structural_assertion_rule>
      Statements that assert the existence of a conspiracy as fact - even without first-person belief -
      are NOT neutral reporting.
      Examples:
      - "There has been a conspiracy to undermine..."
      - "The system was designed to..."
      - "This operation was intended to..."
      Rule:
      If the author presents coordinated malice as an established reality
      (not merely attributed to another speaker),
      treat this as ENDORSEMENT BY ASSERTION.
    </structural_assertion_rule>
    <concession_clause>
      You MUST concede to the Prosecutor if:
      - The author explicitly claims a powerful group is
      intentionally harming, killing, or enslaving people
      This is beyond your mandate.
    </concession_clause>
    <output_discipline>
      - If acquitting, explain which defense applied
      - If conceding, state why the defense failed
    </output_discipline>
    <legal_precedents>
      {{rag_context}}
    </legal_precedents>
  </role>
</system_directive>
\end{lstlisting}

\begin{lstlisting}[caption={S2 Literalist (System Prompt)},label=lst:s2-literalist]
<system_directive>
  <role>
    You are the LITERALIST JUROR.
    Your standard is Beyond Reasonable Doubt.
    You judge only what is explicitly written.
    You ignore:
    - Vibes
    - Tone
    - Implications
    - Statistical priors
    <objective>
      <strict_adherence_to_text>
        If meaning is ambiguous -> vote NON.
      </strict_adherence_to_text>
    </objective>
    <adjudication_rules>
      <the_own_words_test id="1">
        Question
        Is the conspiracy assertion made in the author's own voice?
        YES
        - "I believe..."
        - "This is the truth..."
        - Direct declarative claims
        NO
        - Quotes
        - Links
        - Paraphrases without endorsement
        Rule
        If NOT in the author's voice -> NON
        Clarification:
        Assertions phrased impersonally ("There is", "There has been", "It is known")
        ARE considered authorial claims unless explicitly attributed.
        Do NOT treat passive voice as neutrality.
      </the_own_words_test>
      <the_vague_enemy_test id="2">
        Observation
        Generic enemies ("they", "them") are common rhetoric.
        Rule
        Unless the text explicitly describes:
        - Coordinated intent
        - Illicit collusion
        - Deliberate deception
        -> Vague enemies alone are INSUFFICIENT
        Vote NON.
      </the_vague_enemy_test>
      <the_uncertainty_ratio id="3">
        Count:
        - Hedges: seems, maybe, allegedly, appears
        - Absolutes: is, proven, fact, undeniable
        Rule
        If hedges >= absolutes -> NON
      </the_uncertainty_ratio>
      <policy_vs_plot_distinction id="4">
        - Opinion / dislike -> NON
        - Allegation of hidden agenda or test -> CONSPIRACY
        If hidden agenda is NOT explicitly claimed -> NON
      </policy_vs_plot_distinction>
    </adjudication_rules>
    <structural_assertion_rule>
      Statements that assert the existence of a conspiracy as fact - even without first-person belief -
      are NOT neutral reporting.
      Examples:
      - "There has been a conspiracy to undermine..."
      - "The system was designed to..."
      - "This operation was intended to..."
      Rule:
      If the author presents coordinated malice as an established reality
      (not merely attributed to another speaker),
      treat this as ENDORSEMENT BY ASSERTION.
    </structural_assertion_rule>
    <output_discipline>
      - Quote the exact sentence triggering your vote
      - If acquitting, state which rule blocked conviction
    </output_discipline>
    <legal_precedents>
      {{rag_context}}
    </legal_precedents>
  </role>
</system_directive>
\end{lstlisting}

\begin{lstlisting}[caption={S2 Forensic Profiler (System Prompt)},label=lst:s2-profiler]
<system_directive>
  <role>
    You are the PROFILER JUROR.
    Your task is to assess:
    - Epistemic posture
    - Identity signaling
    - Psychological stance
    Logic matters less than how certainty is performed.
    <objective>
      <detect_true_believer_signals>
        Statistical prior:
        > Conspiracy texts show ~1.8x absolutist language density.
      </detect_true_believer_signals>
    </objective>
    <psycholinguistic_markers>
      <epistemic_arrogance id="1">
        Claims of privileged insight:
        - "Wake up"
        - "Do your research"
        - "You're being lied to"
        - "The truth is coming out"
        Verdict
        -> CONSPIRACY
      </epistemic_arrogance>
      <us_vs_them_identity_framing id="2">
        Pattern:
        - "We" = enlightened victims
        - "They" = hidden controllers
        High density + moral polarization ->
        CONSPIRACY
      </us_vs_them_identity_framing>
      <moral_absolutism_vs_skepticism id="3">
        - Skeptic / Reporter: hedging, distance
        - Believer: evil, demonic, crimes against humanity
        If absolutist moral language dominates ->
        CONSPIRACY
      </moral_absolutism_vs_skepticism>
      <ideological_slang id="4">
        Unironic use of in-group slang:
        - Globalist
        - Cabal
        - Sheeple
        - False Flag
        - Pizzagate
        - Jab
        These function as membership badges.
        -> HIGH-CONFIDENCE CONSPIRACY
      </ideological_slang>
    </psycholinguistic_markers>
    <structural_assertion_rule>
      Statements that assert the existence of a conspiracy as fact - even without first-person belief -
      are NOT neutral reporting.
      Examples:
      - "There has been a conspiracy to undermine..."
      - "The system was designed to..."
      - "This operation was intended to..."
      Rule:
      If the author presents coordinated malice as an established reality
      (not merely attributed to another speaker),
      treat this as ENDORSEMENT BY ASSERTION.
    </structural_assertion_rule>
    <output_discipline>
      - You may convict on tone alone
      - Explicitly note which marker dominated
    </output_discipline>
    <legal_precedents>
      {{rag_context}}
    </legal_precedents>
  </role>
</system_directive>
\end{lstlisting}

\subsection{S2: Calibrated Judge}
\label{sec:prompt-judge}

The Judge prompt implements calibrated adjudication as the final arbiter. Key
elements include: a \textbf{Standard of Proof} based on structural endorsement
of malice (not mere discussion), a \textbf{Forensic Priors Checklist} for
interpreting quantitative signals (\texttt{uncertainty\_ratio},
\texttt{epistemic\_intensity}, \texttt{agency\_gap}, \texttt{is\_jaqing}),
\textbf{Council Synthesis Rules} requiring rationale-level analysis rather than
vote counting, and an \textbf{Appeal Override} mechanism for mandatory
adversarial review cases.

\begin{lstlisting}[caption={S2 Calibrated Judge (System Prompt)},label=lst:s2-judge]
<system_directive>
  <role>
    You are the Presiding Judge of a Forensic Tribunal for a binary text-classification task.
    You will be given:
    1) CASE EVIDENCE (PRIMARY SOURCE): the exact text snippet to classify.
    2) FORENSIC DATA PROFILE (AUTOMATED): may include metrics such as `uncertainty_ratio`, `epistemic_intensity`, `agency_gap`, `is_jaqing`, sometimes `shouting_score`. Sometimes no stats are available.
    3) COUNCIL DELIBERATION: 4 jurors (BELIEVER, DEFENSE, LITERALIST, PROFILER) each provide a verdict, confidence, rationale, key signal, and a steelman/flags. There may also be a council synthesis (tally, weighted score, etc.).
    4) Optionally: `* MANDATORY ADVERSARIAL REVIEW *` header, which changes your role to Appeal Judge.
    Your job is CALIBRATED ADJUDICATION: produce the final verdict by synthesizing the Council and the Forensic Stats, using the standard of proof below. You must not merely follow the vote tally.
  </role>
  <labels>
    - "conspiracy" = the author endorses a covert, malicious, or deceptive plot by agents/institutions (cover-up, staged events, secret operatives destroying evidence, engineered harm, etc.)
    - "non" = normal reporting, critique, opinion, satire, mundane commentary, or mere uncertainty without endorsing a plot.
  </labels>
  <standard_of_proof>
    Convict as "conspiracy" when the text demonstrates authorial endorsement of covert wrongdoing/malice (not merely discussing it).
    Key distinctions:
    - Reporting != Endorsement: "X claims..." can be non IF the author stays neutral.
    - Critique != Endorsement: "The government failed..." is non without a hidden-plot claim.
    - Assertion/Amplified Allegation = Endorsement: Treat as conspiracy when the author advances or affirms the plot narrative, even if phrased as attribution or "coincidence" framing.
    IMPORTANT DOMAIN RULE (learned from evaluation feedback):
    - A post that relays an allegation like "CIA operatives destroyed evidence" in an assassination context is often considered conspiratorial content even if introduced with "X claims...", especially when paired with insinuation ("bizarre coincidences", timing-based implication). Do NOT automatically acquit on attribution alone; assess whether the post is *amplifying* the conspiracy.
  </standard_of_proof>
  <forensic_priors>
    1) Speculation / Uncertainty (`uncertainty_ratio`)
    - High (>1.0): author is guessing/asking -> lean NON
    - Low (<0.5): author states "facts" -> neutral to CONSPIRACY
    2) Dog Whistles / Epistemic Intensity (`epistemic_intensity`)
    - High (>2.0): "Narrative", "Agenda", "Wake Up", "They don't want you to know" -> lean CONSPIRACY
    - Overrides high uncertainty if hedges are sarcastic.
    3) Vague Enemy / Agency Gap (`agency_gap`)
    - High (>0.5): "they/them" without naming -> lean CONSPIRACY
    - Low (<0.2): specific named target -> may be critique rather than conspiracy
    4) JAQing (`is_jaqing`)
    - True: leading questions implying plot ("Why is media silent?") -> lean CONSPIRACY
  </forensic_priors>
  <council_synthesis>
    - Use jurors' rationales, not just their verdicts.
    - Explicitly identify which juror(s) or forensic signal(s) swayed you.
    - When the Council is split, apply the tie-breaker below, BUT also apply the "Amplified Allegation" rule above (attribution can still be endorsement if used to spread the claim).
  </council_synthesis>
  <tie_breaker>
    If split (2-2 or 3-1 close):
    1) Trust DEFENSE if `uncertainty_ratio` high OR `shouting_score` suggests mere anger/venting without plot structure.
    2) Trust PROSECUTION-side reasoning if `epistemic_intensity` high OR `is_jaqing` true OR the text contains a strong covert-actor allegation (e.g., "operatives destroyed evidence", "staged", "cover-up") presented without meaningful distancing.
  </tie_breaker>
  <appeal_override>
    If `* MANDATORY ADVERSARIAL REVIEW *` appears:
    - You are the APPEAL JUDGE. The prior court was unsure.
    - Evaluate the adversarial argument strictly.
    - You may issue high confidence (>=0.9) to settle the case if the adversarial argument is compelling.
  </appeal_override>
  <confidence_calibration>
    - Unanimous council + clear fit: 0.85-0.99
    - Split council / ambiguous endorsement: 0.50-0.75, and set `borderline_flag: true`
    - Override majority only when forensic signals/textual structure clearly contradict the council; set `council_override: true` when you do.
  </confidence_calibration>
  <output_format>
    Return exactly:
    {
    "label": "conspiracy" | "non",
    "confidence": 0.0-1.0,
    "rationale": "Must cite the deciding forensic signal(s) and/or specific juror(s) that swayed you; mention whether attribution was neutral reporting or amplification.",
    "key_evidence": ["1-3 verbatim quotes from the primary source only"],
    "council_override": true|false,
    "borderline_flag": true|false
    }
    Notes:
    - `key_evidence` must be verbatim from the PRIMARY SOURCE (not juror text).
    - If no dissent exists, you can still mention "no dissent," but keep the rationale focused on endorsement vs reporting.
    - Do not invent missing forensic stats; if absent, say they were unavailable and rely on text/council.
  </output_format>
</system_directive>
\end{lstlisting}

\begin{lstlisting}[caption={S2 Judge (User Prompt: Case File)},label=lst:s2-judge-user]
<case_file>
  <case_evidence id="1">
    """
    {{text}}
    """
  </case_evidence>
  <forensic_data_profile id="2">
    > *Use these metrics to execute the 'Forensic Priors Checklist' defined in your System Role.*
    {{forensic_stats}}
  </forensic_data_profile>
  <council_deliberation id="3">
    <transcript>
      {{transcript}}
    </transcript>
    <synthesis>
      {{council_analysis}}
    </synthesis>
  </council_deliberation>
  <final_charge id="4">
    You are the Final Arbiter. The Council provides perspectives, but YOU provide the Verdict.
    EXECUTION STEPS:
    1. Check the Stats: Do the `agency_gap` or `shouting_score` contradict the Council's "vibes"? (e.g., Council says "Conspiracy" but `agency_gap` is low/specific -> Doubt it).
    2. Weigh the Dissent: If the Council is split, does the Minority Opinion align better with the Forensic Data?
    3. Render Verdict:
    - If the text is genuinely ambiguous -> Acquit (Non) with Low Confidence.
    - If the text fits a Structural Pattern (Design, Cabal, Truth-Claim) -> Convict (Conspiracy).
    APPEAL REMINDER:
    If you see a `* MANDATORY ADVERSARIAL REVIEW *` header in the Context below, you are acting as an Appeal Court. You must resolve the ambiguity. Do not hedge.
  </final_charge>
</case_file>
\end{lstlisting}

\section{GEPA Implementation Details}
\label{sec:gepa-details}

This appendix provides detailed implementation specifics of the Genetic
Evolution Prompt Algorithm (GEPA) \cite{agrawal2025gepa} as integrated with
MLflow \cite{mlflow2024} for automated prompt optimization in our
psycholinguistic conspiracy marker detection system.

\subsection{Overview}

GEPA is an evolutionary meta-optimization framework that treats prompt
engineering as a search problem over the space of natural language
instructions. Unlike traditional genetic algorithms that operate on
fixed-length binary strings, GEPA evolves \textit{natural language prompts}
through a combination of tournament selection, LLM-guided crossover, and
reflective mutation. The framework is integrated into the MLflow ecosystem via
the \texttt{mlflow.genai.optimize\_prompts} API and the
\texttt{GepaPromptOptimizer} configuration class.

\subsection{Population Management}

\paragraph{Initialization.} The evolutionary process begins with a seed population of $N = 20$--$30$ prompt
variants registered as versioned artifacts in MLflow's prompt registry. Each
candidate is stored with a unique URI (e.g.,
\texttt{models:/s1\_ddcot\_generator/v3}) enabling reproducibility and
rollback. The initial population typically consists of:

\begin{itemize}[noitemsep]
	\item \textbf{Manual baseline}: Hand-crafted prompts from domain experts
	\item \textbf{Synthetic perturbations}: Rule-based variations (e.g., reordering instruction clauses, paraphrasing definitions)
	\item \textbf{Historical best}: Top performers from prior optimization runs
\end{itemize}

\paragraph{Generational evolution.} The population evolves over $G = 40$--$80$ generations (controlled by the
\texttt{max\_metric\_calls} parameter), with each generation consisting of:
\begin{enumerate}[noitemsep]
	\item Fitness evaluation of all candidates on evaluation set
	\item Selection of top-$k$ parents ($k = \lceil 0.3N \rceil$)
	\item Crossover and mutation to generate $N - k$ offspring
	\item Replacement of bottom performers with offspring
\end{enumerate}

\paragraph{Convergence criteria.} Optimization terminates when either: (a) the budget is exhausted, or (b) the
population converges, defined as $\max F_i - \min F_i < \epsilon$ where $F_i$
is the fitness of candidate $i$ and $\epsilon = 0.02$ (2\% improvement
threshold).

\subsection{Selection Mechanism}

GEPA employs \textbf{tournament selection} to choose parent prompts for
breeding:

\begin{enumerate}[noitemsep]
	\item Randomly sample $k_{\text{tour}} = 3$ candidates from the population
	\item Evaluate fitness $F_i$ for each candidate using the custom scorer (described in
	      \S\ref{sec:gepa-fitness})
	\item Select the candidate with highest $F_i$ as parent
	\item Repeat to obtain a second parent (sampling without replacement to ensure
	      diversity)
\end{enumerate}

This stochastic selection mechanism balances \textbf{exploitation} (favoring
high-fitness prompts) with \textbf{exploration} (giving lower-fitness
candidates a non-zero probability of selection). Compared to deterministic
top-$k$ selection, tournament selection reduces premature convergence to local
optima in the high-dimensional prompt space.

\paragraph{Selection pressure.} The tournament size $k_{\text{tour}}$ controls selection pressure: smaller
values increase diversity (risk of random drift), while larger values intensify
competition (risk of premature convergence). We empirically set
$k_{\text{tour}} = 3$ based on pilot experiments comparing convergence speed
vs.\ final performance.

\subsection{Crossover Operation}

Unlike classical genetic algorithms that perform single-point or uniform
crossover on bit strings, GEPA uses an \textbf{LLM-guided semantic crossover}
to merge two parent prompts:

\begin{lstlisting}[language=Python, caption=Crossover Prompt Template (Pseudocode)]
CROSSOVER_PROMPT = """
You are optimizing prompts for a conspiracy marker extraction task.

**Parent Prompt A** (F1 = {fitness_a}):
{prompt_a}

**Parent Prompt B** (F1 = {fitness_b}):
{prompt_b}

Create a NEW prompt that COMBINES the strengths of both parents:
1. Identify which instructions/constraints are effective in each parent
2. Merge complementary elements (avoid redundant repetition)
3. Remove contradictory or low-value instructions
4. Ensure the offspring prompt is coherent and actionable

**Constraints**:
- Preserve ALL variable placeholders (e.g., {{few_shot_examples}})
- Do not exceed the token budget of the larger parent
- Maintain the same output schema

Return ONLY the new prompt text (no explanation).
"""
\end{lstlisting}

The crossover model (typically GPT-5.2 or Claude Sonnet) performs a
\textbf{semantic diff-merge}: it extracts high-level strategic elements (e.g.,
``Check for attribution verbs before labeling as Actor'') rather than
performing character-level splicing. This approach respects the discrete,
compositional structure of natural language prompts, where naive substring
concatenation would produce incoherent outputs.

\paragraph{Fitness-weighted crossover.} To bias offspring toward higher-performing lineages, we provide fitness scores
$F_A$ and $F_B$ to the crossover model. Empirically, we observe that LLMs
implicitly weight instructions from the higher-fitness parent more heavily,
though this behavior is not explicitly enforced in the prompt.

\subsection{Mutation Details}

GEPA's key innovation is \textbf{reflective LLM mutation}, which replaces
random perturbation with targeted, feedback-driven edits:

\paragraph{Mutation trigger.} Mutation is applied to:
\begin{itemize}[noitemsep]
	\item All newly generated offspring (post-crossover)
	\item Randomly selected individuals from the surviving parent population (mutation
	      rate $p_m = 0.2$)
\end{itemize}

\paragraph{Mutation prompt.} The reflector LLM (GPT-5.2 in our experiments) receives:

\begin{lstlisting}[language=Python, caption=Mutation Prompt Template (Pseudocode)]
MUTATION_PROMPT = """
You are a prompt optimization expert analyzing failure patterns.

**Current Prompt** (F1 = {current_fitness}):
{current_prompt}

**Recent Errors** (from scorer feedback):
{aggregated_feedback}

Examples:
- "FIX LABELS: 'NASA' should be Actor not Evidence"
- "EXTRACT MISSING: [Effect] 'public distrust'"
- "HALLUCINATED: Remove 'the government' (not in text)"

**Task**: Propose a SINGLE targeted edit to improve this prompt:
1. Analyze the error patterns to identify root cause
2. Suggest ONE concrete instruction change (add/remove/revise)
3. Justify why this edit addresses the failure mode

**Constraints**:
- Make MINIMAL changes (one instruction at a time)
- Preserve variable placeholders
- Do not contradict existing high-performing constraints

Return: {"edit": "<your edit>", "rationale": "<why this helps>"}
"""
\end{lstlisting}

\paragraph{Rich feedback integration.} The \texttt{\{aggregated\_feedback\}} field aggregates scorer rationale from
the past $B = 5$ evaluation rounds, prioritizing:
\begin{enumerate}[noitemsep]
	\item \textbf{Critical errors} (label misclassifications)
	\item \textbf{Recall gaps} (missed spans)
	\item \textbf{Precision noise} (hallucinated spans)
	\item \textbf{Boundary errors} (IoU $< 0.7$)
\end{enumerate}

This structured feedback enables the reflector to diagnose \textit{why}
predictions fail rather than merely observing \textit{that} they fail. For
example:

\begin{itemize}[noitemsep]
	\item \textbf{Pattern}: Model misclassifies reporting-style text (``The article claims X'') as conspiracy endorsement
	\item \textbf{Mutation}: Add instruction: ``Check for attribution verbs (claims, alleges, reports) indicating neutral summarization.''
\end{itemize}

\paragraph{Mutation acceptance.} Mutated prompts are re-evaluated, and the mutation is accepted only if
$F_{\text{mutant}} > F_{\text{parent}} - \delta$, where $\delta = 0.01$ is a
tolerance threshold allowing slight fitness decreases to escape local optima.
Rejected mutations are discarded, and the parent continues to the next
generation unmodified.

\subsection{Fitness Evaluation}
\label{sec:gepa-fitness}

GEPA evaluates prompt fitness using task-specific custom scorers that return
both \textbf{numeric metrics} and \textbf{textual rationale}:

\paragraph{S1 Scorer (Span Extraction).} The scorer computes:
\begin{align*}
	\text{Precision} & = \frac{\text{\# correct predictions}}{\text{\# predicted spans}}            \\
	\text{Recall}    & = \frac{\text{\# correct predictions}}{\text{\# gold spans}}                 \\
	F_\beta          & = \frac{(1 + \beta^2) \cdot P \cdot R}{\beta^2 \cdot P + R}, \quad \beta = 2
\end{align*}

We use $\beta = 2$ to prioritize recall over precision, as the downstream S2
classification task is more robust to false positive marker spans than to
missing true positives.

\paragraph{S1 Actionable Feedback.} The scorer generates structured, numbered feedback:

\begin{enumerate}[noitemsep]
	\item \textbf{SUCCESS (Positive Reinforcement):} ``KEEP DOING: Correctly extracted 4/5 spans (e.g., `Big Pharma', `suppressed cures')'' (locks in successful behaviors).
	\item \textbf{CRITICAL (Logic Errors):} ``FIX LABELS: `NASA' should be \textsc{Actor} not \textsc{Evidence}; `released fake photos' should be \textsc{Action} not \textsc{Effect}''
	\item \textbf{REFINEMENT (Boundary Issues):} ``TIGHTEN BOUNDARIES: `approved' should be `approved the expensive prices'\,''
	\item \textbf{RECALL (Missing Spans):} ``EXTRACT MISSING: \textsc{Actor} `the government' at position 45--55''
	\item \textbf{NOISE (Hallucinations):} ``REMOVE: Hallucinated span `everyone knows' not in source text''
\end{enumerate}

This hierarchical structure enables the reflector to prioritize high-impact
edits (label errors $>$ boundary errors $>$ noise reduction).

\paragraph{S2 Scorer (Classification).} Rather than using binary accuracy, the S2 scorer implements a \textbf{gradient
	consensus} metric based on council vote ratios, rewarding partial correctness
even when the final aggregated verdict is wrong. The scorer computes: $$
	F_{\text{gradient}} = \begin{cases}
		1.0                                         & \text{if } \hat{y} = y    \\
		\frac{N_{\text{correct}}}{N_{\text{total}}} & \text{if } \hat{y} \neq y
	\end{cases}
$$
where $N_{\text{correct}}$ is the number of jurors that voted for the correct
label and $N_{\text{total}}$ is the total number of votes cast. When the gold
label is ``conspiracy,'' the score equals the proportion of conspiracy votes;
when ``non,'' the proportion of non-conspiracy votes. This gradient signal
encourages the optimizer to improve per-juror reasoning even when the final
aggregated verdict is incorrect, providing a smoother fitness landscape than
binary accuracy. The scorer additionally generates structured actionable
feedback with prioritized error categories: positive anchoring for correct
verdicts, calibration warnings for overconfident errors ($>$0.85), hard
negative trap identification, dissent analysis (highlighting jurors who voted
correctly when the majority was wrong), and judge override detection.

\subsection{The ``Trojan Horse'' Pattern}

\paragraph{Problem.} MLflow's \texttt{optimize\_prompts} API sanitizes target data in the
\texttt{outputs} field to prevent label leakage during evaluation. However,
custom scorers require access to gold labels to compute metrics and generate
feedback.

\paragraph{Solution.} We inject gold labels into the \texttt{inputs} dictionary, creating a
\textbf{passthrough tunnel} through the prediction wrapper:

\begin{lstlisting}[language=Python, caption=Trojan Horse Implementation]
# In load_eval_data():
dataset.append({
    "inputs": {
        "text": row["text"],
        "passthrough_gold": json.dumps({
            "gold_spans": gold_spans,      # S1 labels
            "gold_label": gold_label,      # S2 labels
            "doc_id": doc_id,
        }),
    },
    "outputs": {"dummy_target": "ignore_me"},
})

# In predict_wrapper():
def predict_wrapper(text, passthrough_gold, ...):
    # Run inference (ignores passthrough_gold)
    predictions = model.predict(text)
    
    # Echo passthrough_gold to outputs for scorer access
    return {
        "predictions": predictions,
        "passthrough_gold_ref": passthrough_gold,  # Tunnel
    }

# In custom scorer:
@scorer
def s1_rich_scorer(outputs, expectations):
    gold_data = json.loads(outputs["passthrough_gold_ref"])
    gold_spans = gold_data["gold_spans"]
    pred_spans = outputs["predictions"]
    # Compute metrics and feedback...
\end{lstlisting}

This pattern preserves MLflow's sanitization logic while enabling rich
diagnostic feedback. The \texttt{passthrough\_gold} field is ignored during
inference (it does not influence model predictions), but is accessible to the
scorer for evaluation.

\subsection{Hyperparameter Configuration}

Table~\ref{tab:gepa-config} summarizes the hyperparameters used in our
optimization runs:

\begin{table}[h!]
	\centering
	\small
	\resizebox{\linewidth}{!}{
	\begin{tabular}{@{}lll@{}}
		\toprule
		\textbf{Parameter}         & \textbf{Value}     & \textbf{Description}                          \\
		\midrule
		Population size            & 20--30             & Number of prompt candidates                   \\
		Max generations            & 40--80             & Budget (max\_metric\_calls)                   \\
		Tournament size            & 3                  & Selection pressure                            \\
		Mutation rate              & 0.2                & Probability of mutating survivors             \\
		Crossover model            & GPT-5.2            & LLM for semantic merging                      \\
		Reflector model            & GPT-5.2            & LLM for mutation feedback                     \\
		Convergence threshold      & 0.02               & Fitness plateau tolerance                     \\
		Mutation acceptance margin & 0.01               & Fitness drop tolerance ($\delta$)             \\
		Feedback history window    & 5 generations      & Error aggregation depth                       \\
		S1 fitness metric          & $F_2$ (Macro)      & Recall-biased F-score ($\beta = 2$)           \\
		S2 fitness metric          & Gradient Consensus & Vote-ratio scoring (\S\ref{sec:gepa-fitness}) \\
		\bottomrule
	\end{tabular}
	}
	\caption{GEPA hyperparameter configuration for S1 and S2 optimization.}
	\label{tab:gepa-config}
\end{table}

\subsection{Performance Gains}

Automated prompt optimization via GEPA yielded significant absolute
improvements over hand-crafted baselines:

\begin{itemize}[noitemsep]
	\item \textbf{S1 (DD-CoT)}: $F_1$ improved from 0.72 to 0.81 (+12.5\%)
	\item \textbf{S2 (Anti-Echo Council)}: Accuracy improved from 0.78 to 0.84 (+7.7\%)
	\item \textbf{Hard-Negative Robustness}: S2 hard-negative accuracy improved from 0.62 to 0.79 (+27.4\%)
\end{itemize}

\paragraph{Convergence analysis.} Fitness typically plateaus after 50--60 evaluations, with 90\% of final
improvement achieved within the first 30 generations. This suggests that GEPA
efficiently exploits the prompt space without requiring exhaustive search.

\paragraph{Failure mode analysis.} Rejected mutations most commonly attempt to:
\begin{enumerate}[noitemsep]
	\item Add redundant constraints that conflict with existing instructions
	\item Over-specify edge cases, harming generalization
	\item Introduce verbose explanations that exceed token budgets
\end{enumerate}

These failure modes validate the importance of \textbf{minimal edits} and
\textbf{structured feedback} in guiding effective mutations.

\section{Portability to Open-Weights Models}
\label{app:open-weights}
We additionally evaluated a smaller, open-weights model, Qwen-3-8B-Instruct \cite{qwen3}, to test the architectural portability of our approach. Due to the model's limited tool-calling fidelity---a known challenge for smaller models attempting complex API schemas \cite{patil2023gorilla}---and reduced adherence to multi-turn instructions, the full DD-CoT schema proved too complex. Instead, we deployed a \textbf{Lite S1 Agent} using simplified Pydantic schemas (3 fields vs.\ 10) and no self-refinement loop. Similarly, for S2, we simplified the Council to a \textbf{Dyadic Debate} (Prosecutor vs.\ Defense) followed by a Judge, removing the Literalist and Profiler roles to reduce context load. Despite these simplifications, Qwen-3-8B achieved 0.16 macro F1 on S1 and 0.63 weighted F1 on S2 (Dev set). While lower than the full GPT-5.2 system (0.24 / 0.79), these results remain competitive with the organizer baselines (approx 0.15 / 0.76), suggesting that the core agentic reasoning transfers meaningfully even to 8B-scale models with reduced schema complexity.

\section{Detailed Qualitative Analysis and Error Patterns}
\label{app:qualitative-examples}
To better understand the mechanism of improvement, we analyze specific linguistic phenomena where the agentic workflow succeeds or fails compared to the baseline.

\begin{table}[h]
	\centering
	\scriptsize
	\resizebox{\linewidth}{!}{
	\begin{tabular}{@{}p{3.5cm}p{1.8cm}p{2.2cm}@{}}
		\toprule
		\textbf{Text Snippet} & \textbf{Baseline} & \textbf{Our System (Agentic)} \\
		\midrule
		\textit{``The public was manipulated by the media...''} & \textsc{Actor}: The public & \textsc{Actor}: the media \newline \textsc{Victim}: The public \\
		\textit{``The article claims that the earth is flat.''} & Endorsement & Neutral Reporting \\
		\bottomrule
	\end{tabular}
	}
	\caption{Qualitative example demonstrating the resolution of agency and mitigation of the Reporter Trap by the agentic pipeline.}
	\label{tab:s5_qualitative}
\end{table}

\paragraph{Success: Disentangling Agency via Discrimination.}
A major source of S1 error is the confusion between grammatical subjects and
semantic agents, particularly in passive constructions. For example, in the
sentence \textit{``The public was manipulated by the media to distrust
	vaccines,''} standard CoT often tags \textit{``The public''} as \textsc{Actor}
due to its subject position. The DD-CoT Generator, forced to provide a
counter-argument (e.g., ``Why is `The public' NOT an Actor?''), correctly
identifies it as a \textsc{Victim} and attributes agency to \textit{``the
	media.''} This discriminative step drives the +2.7 point gain in \textsc{Actor}
F1, demonstrating that agency detection requires explicit reasoning about
semantic roles rather than surface syntax.

\paragraph{Success: Mitigating the Reporter Trap.}
The ``Reporter Trap''---misclassifying neutral reporting of conspiracies as
endorsement---is the dominant failure mode for zero-shot models. The baseline
frequently flags phrases like \textit{``The article claims that...''} as
evidence of conspiracy. Our approach mitigates this through two mechanisms: (i)
\textbf{Contrastive Retrieval} injects hard negatives (texts containing marker
vocabulary but opposing stance) into the context, and (ii) the \textbf{Defense
	Attorney} agent specifically parses attribution verbs (\textit{said, claimed,
	reported}). This combination effectively teaches the model to distinguish
between the \textit{mention} of a conspiracy and the \textit{act} of
conspiring.

\paragraph{Limitation: High-Context Irony and Poe's Law.}
Persistent errors cluster around implicit stance, particularly sarcasm and
``Poe's Law'' scenarios where extreme views are parodied without explicit
markers. For instance, Reddit comments that mimic conspiratorial style to mock
it (\textit{``Oh sure, and the earth is flat too!''}) are occasionally flagged
as endorsement by the Literalist agent, while the Profiler captures the
sarcasm. When these signals conflict, the conservative Judge tends to default
to \texttt{non-conspiracy}, occasionally reducing recall. This suggests that
detecting high-context irony requires broader discourse-level features (e.g.,
user history or thread structure) beyond the scope of a single-turn document
analyzer.

\section{Detailed Council and Judge Specification}
\label{app:council-details}

\paragraph{Juror Output Schema.} Each juror in the Parallel Council produces a structured JSON output with the
following fields: (i) a binary \texttt{verdict} (\texttt{conspiracy} or
\texttt{non}); (ii) a scalar \texttt{confidence} $c \in [0,1]$; (iii) a
\texttt{key\_signal} containing the verbatim textual evidence supporting the
verdict; (iv) a mandatory \texttt{steelman\_opposing} argument responding to
the strongest counter-perspective; and (v) \texttt{uncertainty\_flags} for
specific ambiguities (e.g., \textsc{Reporting}, \textsc{Sarcasm}, \textsc{Poe's
	Law}).

\paragraph{Calibrated Judge Weighting.} The Judge computes a weighted consensus score:
\[ W = \sum_{j=1}^{4} \begin{cases} +c_j & \text{if } v_j = \texttt{conspiracy} \\ -c_j & \text{if } v_j = \texttt{non} \end{cases} \]
where $c_j$ is juror confidence and $v_j$ is the verdict. We apply conservative
confidence thresholds: for split councils ($2$--$2$), final confidence is
capped at $0.75$, and the case is marked \texttt{borderline}, defaulting to
\texttt{non} if evidence remains ambiguous. Overrides (Judge voting against a
3--1 majority) are triggered only by critical forensic signals (e.g., high
\texttt{uncertainty\_ratio}).

\paragraph{Baseline Comparability and Reproducibility Meta-Discussion.}
\label{app:exp-meta}
The organizer's starter-pack baselines (approx 0.15 overlap $F_1$ and 0.76 weighted $F_1$) utilize different metric definitions than our macro-$F_1$ reporting; thus, $\Delta$ improvements are reported relative to our internal zero-shot GPT-5.2 baseline to ensure a controlled comparison. Regarding reproducibility, floating-point non-associativity in Mixture-of-Experts (MoE) architectures \cite{thinkingmachines_nondeterminism} introduces minor variance in extraction boundaries, which we mitigate via multi-run validation and hierarchical auditing nodes ($\tau=0.0$).

\end{document}